\documentclass[10pt,twocolumn,letterpaper]{article}
\let\arxivArticleSection\section
\let\arxivArticleSubsection\subsection
\let\arxivArticleSubsubsection\subsubsection
\usepackage[pagenumbers]{cvpr}
\usepackage[T1]{fontenc}
\usepackage{array,multirow,colortbl}
\usepackage{placeins,float}
\usepackage[pass]{geometry}
\graphicspath{{figures/}}
\newcommand{\ourrow}{\rowcolor{gray!12}}
\definecolor{cvprblue}{rgb}{0.21,0.49,0.74}
\usepackage[breaklinks,colorlinks,allcolors=cvprblue]{hyperref}
\hypersetup{
  bookmarksnumbered=true,
  pdftitle={GeoVLDrive: Geometry-Grounded Vision-Language Reasoning for Autonomous Driving}
}
\title{GeoVLDrive: Geometry-Grounded\\
  Vision-Language Reasoning for Autonomous Driving}
\author{%
\normalsize
\textbf{JiaHui Huang}$^{1}$ \quad
\textbf{Songyan Zhang}$^{2}$ \quad
\textbf{Hanbing Li}$^{1}$ \quad
\textbf{Hanchao Leng}$^{1}$ \\
\normalsize\textbf{Kuiyuan Yang}$^{1}$ \quad
\textbf{Guang Chen}$^{1,3}$ \quad
\textbf{Hangjun Ye}$^{1}$ \quad
\textbf{Chen Lv}$^{2}$ \\
\normalfont\normalsize $^{1}$Xiaomi Corporation \quad
$^{2}$Nanyang Technological University \quad
$^{3}$University of Missouri
}
\hypersetup{pdfauthor={JiaHui Huang, Songyan Zhang, Hanbing Li, Hanchao Leng, Kuiyuan Yang, Guang Chen, Hangjun Ye, Chen Lv}}

\newcolumntype{P}[1]{>{\raggedright\arraybackslash}p{#1}}
\newcommand{\supplementtableformat}{%
  \centering
  \setlength{\abovecaptionskip}{0pt}%
  \setlength{\belowcaptionskip}{6pt}%
}
\newcommand{\continuedfigurecaption}[2]{%
  \caption*{Figure~\ref{#1} (continued): #2}%
}
\makeatletter
\newcommand{\arxivSupplement}{%
  \clearpage
  \onecolumn
  \newgeometry{left=1in,right=1in,top=.85in,bottom=.90in,twocolumn=false}%
  \let\section\arxivArticleSection
  \let\subsection\arxivArticleSubsection
  \let\subsubsection\arxivArticleSubsubsection
  \captionsetup{font=normalsize,labelfont=normalfont,textfont=normalfont,labelsep=colon,skip=10pt}%
  \captionsetup[table]{position=top,skip=6pt}%
  \renewcommand{\textfraction}{.2}%
  \renewcommand{\floatpagefraction}{.5}%
  \renewcommand{\topfraction}{.7}%
  \renewcommand{\bottomfraction}{.3}%
  \renewcommand{\dblfloatpagefraction}{.5}%
  \renewcommand{\dbltopfraction}{.7}%
  \setcounter{topnumber}{2}%
  \setcounter{bottomnumber}{1}%
  \setcounter{totalnumber}{3}%
  \setlength{\parindent}{15pt}%
  \setlength{\@fptop}{0pt}%
  \setlength{\@fpsep}{\floatsep}%
  \setlength{\@fpbot}{0pt plus 1fil}%
  \raggedbottom
  \setlength{\emergencystretch}{1.5em}%
  \appendix
  \setcounter{figure}{0}%
  \setcounter{table}{0}%
  \setcounter{equation}{0}%
  \renewcommand{\thefigure}{S\arabic{figure}}%
  \renewcommand{\thetable}{S\arabic{table}}%
  \renewcommand{\theequation}{S\arabic{equation}}%
  \renewcommand{\theHfigure}{supp.\arabic{figure}}%
  \renewcommand{\theHtable}{supp.\arabic{table}}%
  \renewcommand{\theHequation}{supp.\arabic{equation}}%
  \pdfbookmark[0]{Supplementary Material}{supplement}%
  \begin{center}
    {\Large\bfseries GeoVLDrive: Geometry-Grounded Vision-Language\\
     Reasoning for Autonomous Driving\par}
    \medskip
    {\large\bfseries Supplementary Material\par}
  \end{center}
}
\makeatother

\date{}

\begin{document}
\maketitle
\begin{abstract}
\frenchspacing\hyphenpenalty=1600\relax
Vision-language-action (VLA) planners inherit semantic knowledge from
pretrained vision-language models, but their reasoning is rarely grounded in
3D scene structure and its evolution over time. We introduce
\textbf{GeoVLDrive}, a geometry-grounded VLA framework for 3D spatiotemporal
reasoning and planning. Multi-layer features from a 3D foundation model augment
the vision-language backbone through gated cross-attention.
A structured latent chain of thought (CoT) organizes future
scene geometry, key-agent motion, and meta action into ordered, separately
supervised states. An action expert generates trajectory proposals from these
states and multi-depth scene features, and ranks them with scene-grounded,
interacting score tokens. We also introduce \textbf{NAVSIM~3DQA} to train and
evaluate metric 3D spatial and motion understanding.
GeoVLDrive achieves \textbf{$\mathbf{91.8}$ PDMS} on NAVSIM~v1 and \textbf{$\mathbf{90.9}$ EPDMS}
on NAVSIM~v2. On NAVSIM~3DQA, the geometry-grounded backbone scores $89.2$\% overall,
$4.6$ percentage points above the fine-tuned Qwen3-VL-2B baseline.
On \mbox{LingoQA}, it reaches $68.6$\% Lingo-Judge accuracy.\par
\end{abstract}

\begin{figure*}[t]
  \centering

  \includegraphics[width=.88\textwidth]{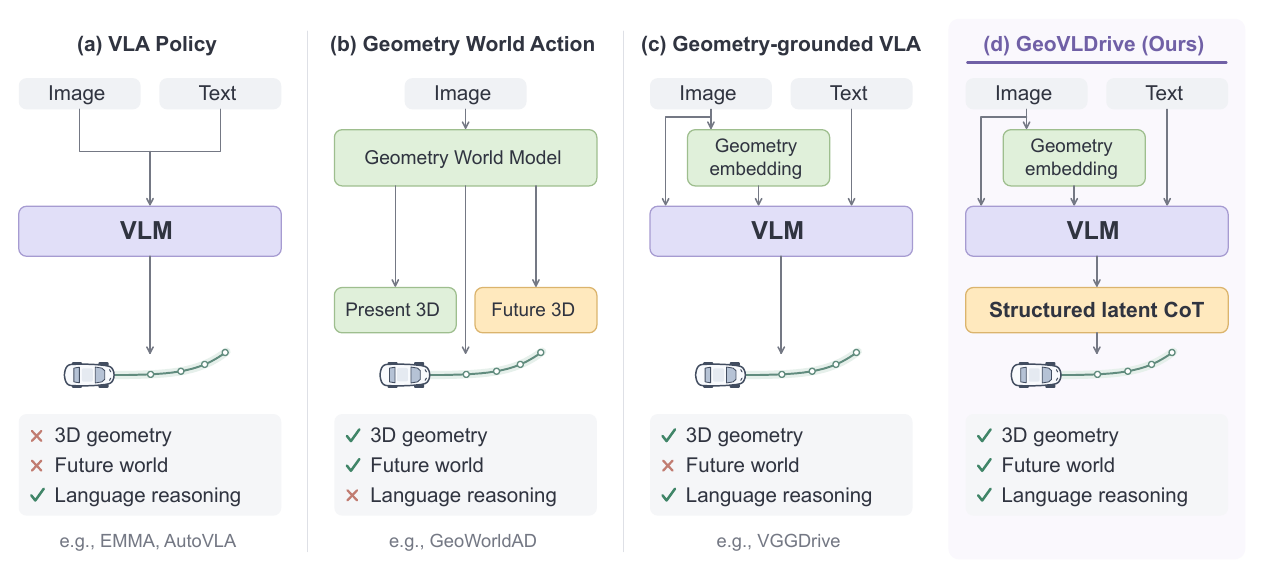}
  \caption{\textbf{Comparison of representative driving frameworks.}
  (a)~VLA policies map images and text to trajectories through a VLM
  \cite{emma,autovla}. (b)~Geometry world action models condition planning
  on observed and predicted geometric representations \cite{geoworldad}.
  (c)~Geometry-grounded VLAs augment VLMs with observed-scene geometric
  features \cite{vggdrive}. (d)~GeoVLDrive combines geometric grounding
  with a structured latent CoT to reason about future scene evolution and
  guide trajectory planning.}
  \label{fig:teaser}
\end{figure*}

\section{Introduction}
\label{sec:intro}

\begingroup\frenchspacing
End-to-end autonomous driving has advanced through joint optimization of
perception, prediction, and planning~\cite{st-p3,uniad,vad,vadv2}. Building on
the semantic knowledge of pretrained vision-language models,
vision-language-action (VLA) frameworks use visual observations and language
inputs to support driving trajectory generation~\cite{orion,emma,drivevlm,opendrivevla,autovla}.
Driving requires reasoning about traffic participants, scene geometry, and
future scene evolution~\cite{uniad,vadv2,transfuser,navsim}. This motivates
grounding such reasoning in 3D geometry and scene semantics to connect
current spatial relationships and their future evolution with ego planning.

Representative approaches differ in how they incorporate semantic reasoning and
geometric representations into trajectory planning (Fig.~\ref{fig:teaser}).
VLA policies (Fig.~\ref{fig:teaser}(a)) build on pretrained vision-language
models for scene understanding and reasoning to guide trajectory
generation~\cite{orion,emma,opendrivevla}. Geometry world action models
(Fig.~\ref{fig:teaser}(b)) encode current scene geometry and predict its future
evolution, conditioning planning on geometric representations of the observed
and anticipated scene~\cite{geoworldad}. Geometry-grounded VLAs
(Fig.~\ref{fig:teaser}(c)) use observed-scene 3D geometric features or reconstruction
supervision to support spatial understanding and trajectory
planning~\cite{vggdrive,vlga}.
\par\endgroup

A central challenge is to connect scene-wide geometric forecasts with
agent-specific motion reasoning and ego maneuvers. Recent VLAs incorporate
geometric and dynamic knowledge into latent representations used for
planning~\cite{lastvla,coworldvla,driveworldvla}. We focus on structuring
this connection through distinct, separately supervised latent groups for
future scene geometry, the motion of a selected key agent, and meta action.
Directed dependencies link scene forecasts to agent motion and maneuver selection.

We propose \textbf{GeoVLDrive}, a
geometry-grounded VLA framework for 3D spatiotemporal
reasoning and planning (Fig.~\ref{fig:teaser}(d)). Multi-layer geometric
features augment the vision-language backbone with 3D spatial information.
A structured latent chain of thought (CoT) organizes future scene geometry,
key-agent motion, and meta action into separately supervised latent states.
Its attention mask lets future-motion states read future-geometry states and meta-action
states read both. An action expert uses the latent CoT and \mbox{multi-depth} geometric
and semantic features to jointly generate and score trajectory proposals. Dedicated score
tokens incorporate scene and reasoning context and exchange information across
proposals, allowing each candidate to be evaluated in relation to the
alternatives. We construct \textbf{NAVSIM~3DQA} to train and evaluate
box-conditioned metric 3D and motion understanding on NAVSIM scenes.

Our main contributions are summarized as follows:
\begin{itemize}
  \item We propose \textbf{GeoVLDrive}, a geometry-grounded VLA framework
  with a structured latent CoT that connects future scene geometry,
  key-agent motion, and meta action through separately supervised latent
  states. With the same action-expert architecture, the latent CoT and
  geometry injection raise NAVSIM~v1 PDMS from $90.1$ to $91.8$.
  \item We develop an \textbf{action expert} that jointly generates and
  scores trajectory proposals conditioned on the latent CoT and multi-depth
  scene features. Dedicated score tokens attend to scene context and to one
  another. This scoring design improves PDMS by $1.6$ over an MLP readout
  of max-pooled action states.
  \item We construct \textbf{NAVSIM~3DQA}, a dataset of $1.28$M
  box-conditioned questions on object category, metric position, vehicle
  dimensions, and motion in NAVSIM scenes. With geometry injection and
  spatial alignment, our backbone scores $89.2$\% overall, $4.6$ percentage
  points above the fine-tuned Qwen3-VL-2B baseline.
\end{itemize}

\begin{figure*}[t]
  \centering

  \includegraphics[width=\textwidth]{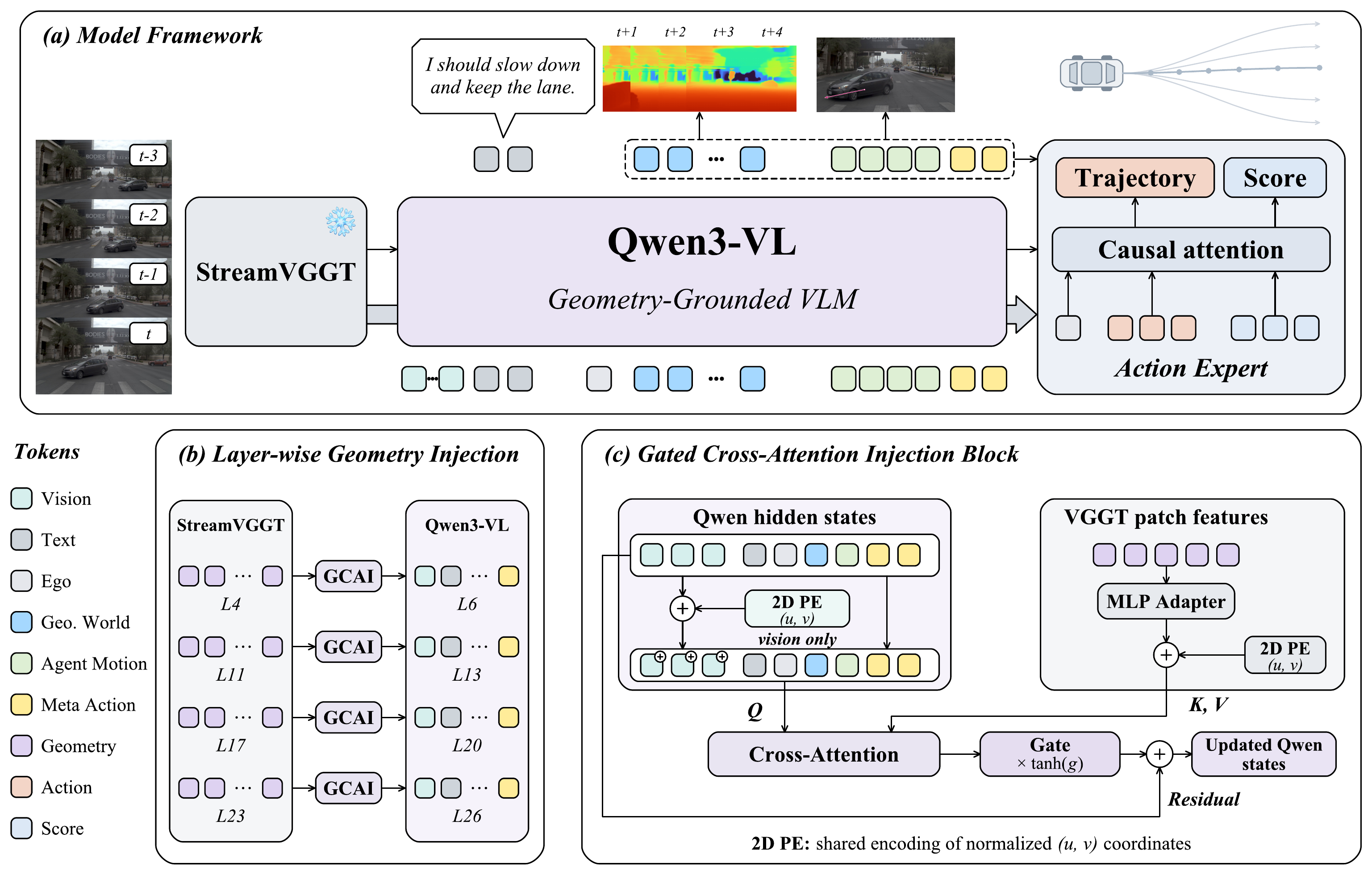}

\caption{\textbf{Overview of GeoVLDrive.}
  \textbf{(a) Model framework.} The geometry-grounded VLM encodes future
  scene geometry, key-agent motion, and meta action in a structured latent CoT that
  conditions the action expert for joint trajectory generation and scoring.
  \textbf{(b) Layer-wise geometry injection.} Gated cross-attention injection
  (GCAI) blocks inject features from multiple StreamVGGT layers at selected
  VLM decoder depths.
  \textbf{(c) Gated cross-attention injection block.} Gated cross-attention injects
  geometry into VLM states, with shared 2D positional encoding for spatial
  alignment.}
  \label{fig:framework}
\end{figure*}

\section{Related Work}
\subsection{Vision-Language-Action Models for Driving}
\begingroup\frenchspacing\doublehyphendemerits=100000\relax
VLA models use pretrained semantic knowledge for trajectory planning.
DriveLM~\cite{drivelm} frames driving understanding as
graph-structured question answering, and EMMA~\cite{emma} incorporates textual
reasoning into action generation. Textual chains of thought~\cite{cot} expose
intermediate reasoning but add autoregressive decoding overhead.
\mbox{AutoVLA}~\cite{autovla} and AdaThinkDrive~\cite{adathinkdrive} adapt the use of
such reasoning to driving scenarios. Geometry-grounded models add 3D information.
VGGDrive~\cite{vggdrive} injects features from a frozen geometry backbone
via multi-layer cross-attention. VLGA~\cite{vlga} supervises a geometry expert
with dense pointmap reconstruction, and DVGT-2~\cite{dvgt2} jointly reconstructs
current geometry and predicts trajectories. Selection-based planners, including
Hydra-MDP~\cite{hydramdp}, WoTE~\cite{wote}, and
iPad~\cite{ipad}, use learned scores to select candidate trajectories. GeoVLDrive
combines geometric grounding with structured reasoning about future scene evolution
for ego planning. It uses iPad's proposal formulation with dedicated score tokens.
\par\endgroup

\subsection{Latent Reasoning and World Modeling}
\begingroup\doublehyphendemerits=100000\relax
Latent CoT methods replace explicit reasoning traces with continuous intermediate
states. Coconut~\cite{coconut} recurrently propagates hidden states,
CCoT~\cite{ccot} learns compressed traces, CODI~\cite{codi} uses hidden-state
self-distillation, and SIM-CoT~\cite{simcot} adds step-level supervision through an
auxiliary decoder. In driving, LaST-VLA~\cite{lastvla} and
CoWorld-VLA~\cite{coworldvla} transfer geometric and dynamic priors into separately
supervised latent groups. PerceptDrive~\cite{perceptdrive} routes distilled
perception priors into world-action modeling; DriveWorld-VLA~\cite{driveworldvla}
shares latent features between world modeling and planning.
OneVL~\cite{onevl} uses language and future-frame decoders to supervise
shared latents. DynVLA~\cite{dynvla} predicts compact dynamics tokens,
LCDrive~\cite{lcdrive} interleaves action proposals with predicted outcomes, and
ColaVLA~\cite{colavla} derives meta-action representations for parallel planning.
Other approaches support planning through video
prediction~\cite{vista,drivedreamer,drivinggpt}, latent forecasting~\cite{law},
occupancy~\cite{occworld}, occupancy-flow prediction~\cite{drivepi}, and future
geometry prediction~\cite{geoworldad}. \mbox{SGDrive}~\cite{sgdrive} is
closest to our design. It supervises scene, agent, and goal queries in a VLM
with occupancy, safety-critical agents, and a $4$-s goal pose, learning geometry
through occupancy supervision, and a block-wise mask keeps the three groups
from attending to one another. GeoVLDrive instead injects 3D foundation-model
features into the VLM, predicts future depth from residually updated geometric
features, and supervises at most one planning-relevant key agent and meta
actions. Its attention mask lets future-motion states read future-geometry states and
meta-action states read both, and blocks the reverse direction. The planner
uses these states to generate and score trajectory proposals.
\par\endgroup

\section{Method}
\label{sec:method}

GeoVLDrive grounds a vision-language backbone in 3D geometry and introduces a
structured latent CoT connecting future scene geometry, key-agent motion, and meta
action (Fig.~\ref{fig:framework}). An action expert jointly generates and scores
trajectory proposals conditioned on the latent CoT and multi-depth geometric
and semantic features.

\subsection{Geometry-Grounded Vision-Language Backbone}
\label{sec:geo}

We combine a pretrained Qwen3-VL-2B-Instruct backbone~\cite{qwen3vl} with a
frozen StreamVGGT geometry encoder~\cite{streamvggt,geoworldad}. The VLM encodes
the visual input, driving prompt, and current ego state, while StreamVGGT
aggregates four front-camera frames to extract current-frame geometric features
$\{G^\ell\}_{\ell}$ from multiple layers.

\paragraph{Layer-wise gated cross-attention for geometry injection.}
We inject geometry features at multiple VLM depths~\cite{vggdrive}, pairing
StreamVGGT decoder layers $[4, 11, 17, 23]$ with Qwen3-VL decoder layers
$[6, 13, 20, 26]$, using zero-based layer indices
(Fig.~\ref{fig:framework}(b)). For each paired depth,
the decoder hidden states query the projected geometry tokens through a gated
cross-attention residual~\cite{flamingo}:
\begin{equation}
\begin{aligned}
X'_m &= X_m + \tanh(g_m)\, A_m, \\
A_m &=
\operatorname{Attn}\!\left(
\operatorname{LN}(X_m),\,
\phi_m(G^\ell),\,
\phi_m(G^\ell)
\right).\end{aligned}
\label{eq:inject}
\end{equation}
where $m$ and $\ell$ index the paired VLM and geometry layers.
$X_m$ denotes the output of VLM decoder layer $m$ before geometry injection,
and $X'_m$ denotes the updated states.

$\operatorname{LN}$ denotes layer normalization, and $\operatorname{Attn}$
denotes query-key-value attention. The projection $\phi_m$ maps geometry features
to the decoder width, and $g_m$ is
a learnable scalar gate that controls the amount of geometric information added
at each depth. Because the queries are formed from the decoder hidden states,
visual, textual, and latent reasoning tokens can all access the geometry stream.
Eq.~\eqref{eq:inject} omits the positional encodings described below.

\paragraph{Spatial alignment of semantic and geometric features.}
To spatially align the VLM's visual representations with geometry features,
we encode normalized image coordinates $(u,v)$ using a shared 2D sinusoidal
positional encoding. The resulting encodings are added to the visual queries and
the geometry keys and values (Fig.~\ref{fig:framework}(c)).
This provides a common image-plane reference across feature grids of
different resolutions, helping cross-attention associate features at
corresponding locations. Together, geometry injection and spatial encoding improve the
backbone's 3D scene understanding and metric localization, as shown by the
NAVSIM~3DQA ablations in Table~\ref{tab:ablation_geo}.

\subsection{Structured Latent Chain-of-Thought}
\label{sec:chain}

We organize future scene geometry, key-agent motion, and meta action into three
supervised groups of latent states that form the latent CoT. The annotations
select at most one key agent per scene (Sec.~\ref{sec:app_chain} of the supplementary material). At
inference, these states are computed in one backbone forward pass. During training, textual CoT provides autoregressive language supervision.

We predict scene geometry over $2$\,s and key-agent motion over
$4$\,s, using residual representations relative to the current state.
The action expert reads both these hidden states and scene features directly
(Sec.~\ref{sec:expert}).

\begin{figure}[t]
  \centering
  \includegraphics[width=\linewidth]{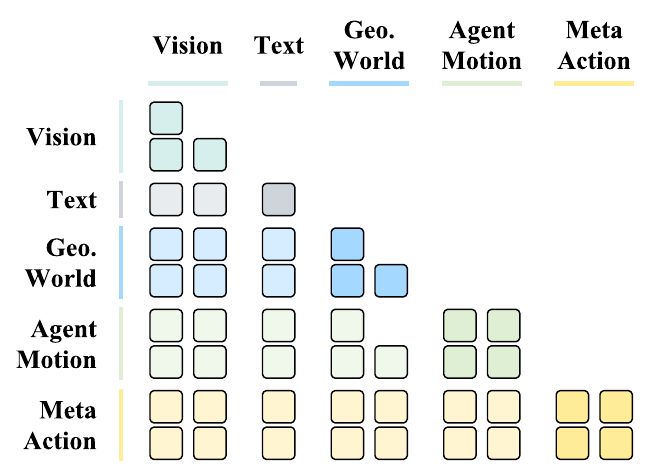}
  \caption{\textbf{Latent CoT attention mask.} Rows denote queries and columns
  denote keys; colored cells indicate allowed direct attention. The motion block shows example future-motion states.}
  \label{fig:mask}
\end{figure}

\paragraph{Structured attention mask.}
The mask in Fig.~\ref{fig:mask} orders the latent CoT.
Future key-agent motion states attend to future geometry states.
Meta-action tokens attend to both groups. The mask blocks reverse connections
between groups. Geometry tokens attend to their own and earlier prediction horizons.
Future-motion tokens attend to one another bidirectionally. Their direct
access to geometry states is limited to the same or earlier horizons.

\paragraph{Future scene geometry.}
We retain the current multi-layer geometric features $\{G^\ell\}_{\ell}$ as a
dense spatial reference, allowing compact latent tokens to focus on temporal
changes. For each prediction horizon $t_k=k\delta$, $k=1,\ldots,K$, a
time-conditioned residual decoder updates these features using $32$ latent
states $\mathbf{F}_k$, the VLM hidden states at the future-geometry tokens of
horizon $k$:
\begin{equation}
\begin{aligned}\Delta G_k^\ell &= \mathcal{R}(G^\ell,\mathbf{F}_k,t_k),\\
(\hat{\mathbf{z}}_k,\mathbf{c}_k) &=
\mathcal{D}_{\mathrm{geo}}
\left(
\left\{G^\ell+\Delta G_k^\ell\right\}_{\ell}
\right),\end{aligned}
\label{eq:wm}
\end{equation}
where $\mathcal{R}$ denotes the decoder's net feature update and
$\mathcal{D}_{\mathrm{geo}}$ produces the future depth and confidence maps
$\hat{\mathbf{z}}_k$ and $\mathbf{c}_k$. On valid future-depth pixels, we use a
loss adapted from VGGT and GeoWorldAD~\cite{vggt,geoworldad}:
\begin{equation}
\mathcal{L}_{\mathrm{world}}
=
\sum_{k=1}^{K}
\left(
\mathcal{L}_{\mathrm{depth}}^{(k)}
+
\mathcal{L}_{\mathrm{conf}}^{(k)}
+
\mathcal{L}_{\mathrm{grad}}^{(k)}
\right),
\label{eq:world_loss}
\end{equation}
combining depth regression, confidence-weighted regression, and multi-scale
gradient consistency (Sec.~\ref{sec:app_losses} of the supplementary material).

\begin{figure*}[t]
  \centering
  \includegraphics[width=\textwidth]{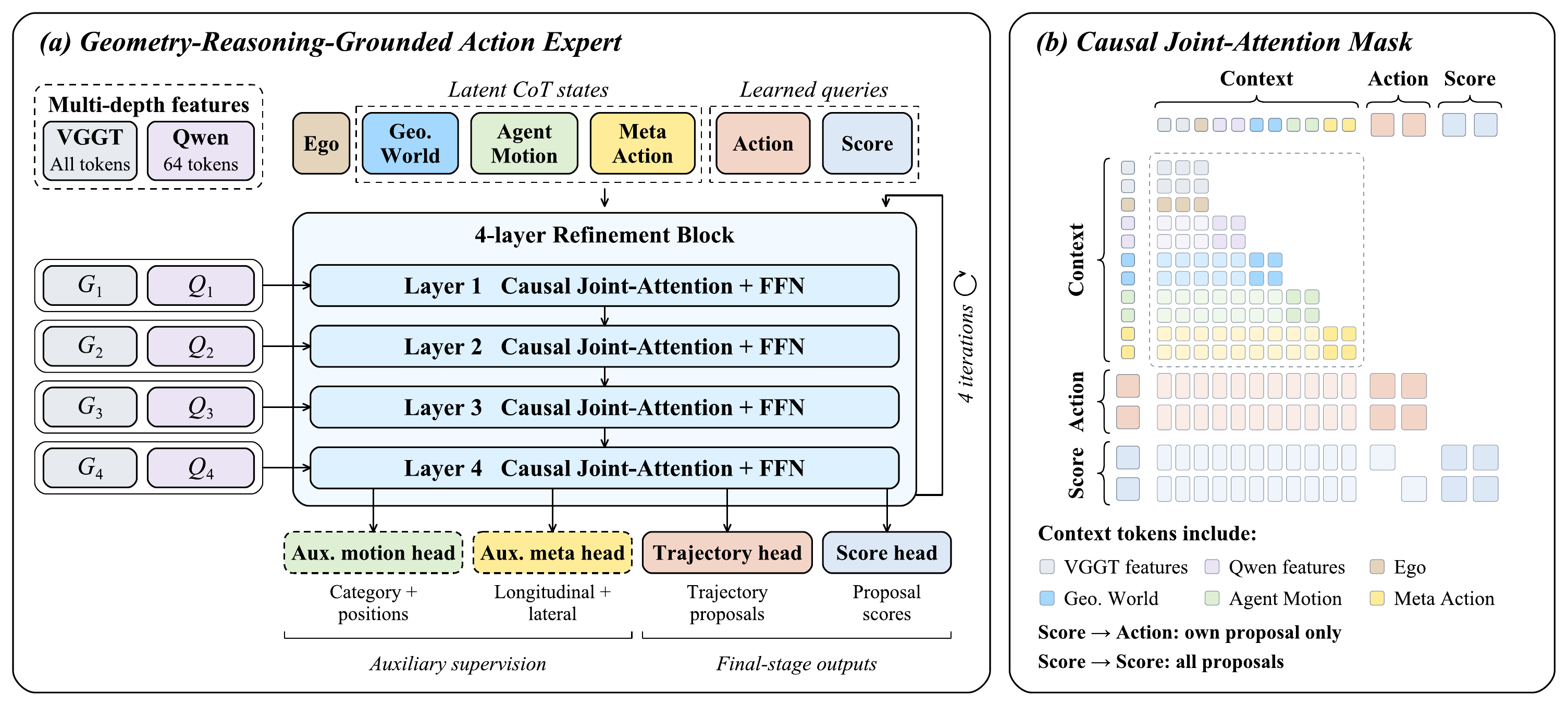}
  \caption{\textbf{Action expert.}
  (a) Overall architecture for joint trajectory generation and scoring,
  integrating multi-depth geometric ($G_i$) and semantic ($Q_i$) features with
  the latent reasoning chain. (b) Causal joint-attention mask used in the
  joint-attention layers of (a).}
  \label{fig:action_expert}
\end{figure*}

\paragraph{Key-agent motion.}
We represent a planning-relevant agent with latent states
$\mathbf{D}=[\mathbf{d}_c;\mathbf{d}_0;\mathbf{d}_{1:J}]$, encoding its category,
current position, and future position offsets. Category and current-position
states attend to the observed context, while future-motion states additionally
attend to future-geometry states. Prediction extends to $J>K$ steps on the same
temporal grid, $t_j=j\delta$, with future positions reconstructed relative to
the current position:
\begin{equation}
  \hat{\mathbf{p}}_0=g_0(\mathbf{d}_0),\qquad
  \hat{\mathbf{p}}_j=\hat{\mathbf{p}}_0+g_\Delta(\mathbf{d}_j),
  \quad j=1,\ldots,J,
  \label{eq:dyn}
\end{equation}
where $g_0$ estimates the current position and $g_\Delta$ predicts offsets from
it. Category and track annotations supervise the predictions, including a
``none'' category for scenes without a key agent:
\begin{equation}
  \mathcal{L}_{\mathrm{agent}}
  = \lambda_{\mathrm{cls}}\mathcal{L}_{\mathrm{CE}}
  + \lambda_{\mathrm{pos}}\mathcal{L}_{\mathrm{SmoothL1}}
  + \lambda_{\mathrm{dir}}\mathcal{L}_{\mathrm{cos}}.
  \label{eq:dyn_loss}
\end{equation}
Smooth L1 regression averages losses on the current position, future offsets,
and reconstructed positions. The cosine loss aligns consecutive-step
displacement directions, with reduced weights for small ground-truth
displacements. Position and direction losses apply only to valid agent tracks.

\paragraph{Meta action.}
Two meta-action tokens represent longitudinal and lateral intent using the
observed context, future geometry states, and key-agent motion states.
Their maneuver distributions $\boldsymbol{\pi}_d$ are supervised by
the corresponding labels $y_d$:
\begin{equation}
  \mathcal{L}_{\mathrm{meta}}
  = \sum_{d\in\{\mathrm{lon},\,\mathrm{lat}\}}
  \operatorname{CE}(\boldsymbol{\pi}_d,y_d).
  \label{eq:meta_loss}
\end{equation}

\subsection{Action Expert}
\label{sec:expert}

Our action expert jointly generates and scores trajectory proposals conditioned
on the latent chain and multi-depth geometric and semantic features
(Fig.~\ref{fig:action_expert}). Causal joint attention directs information from
reasoning to actions and from actions to scores. Dedicated score tokens combine scene context with proposal-specific action states
and exchange information across candidates for trajectory selection.

\paragraph{Multi-depth joint attention.}
The expert is a dedicated transformer operating on latent reasoning states,
ego-state tokens, $R\times J$ action tokens, and $R$ score tokens. Geometric and
semantic features from multiple depths are concatenated with these states at
corresponding expert layers. At each selected depth, Qwen features are compressed
to $64$ tokens by attention pooling with learned queries. All VLM-derived
inputs are detached before the trainable pooling and projection layers,
preventing planning gradients from updating the VLM. Masked self-attention and an
FFN jointly update the resulting sequence under the causal joint-attention mask in
Fig.~\ref{fig:action_expert}(b), which enforces the token-group visibility order
while allowing bidirectional attention within each group.
The refined reasoning, action, and score states
propagate across $S$ refinement stages, with stage $s$ decoding $R$ trajectory
proposals $\boldsymbol{\tau}^{(s)}_r$, each comprising $J$ poses. Trajectory
supervision follows the winner-takes-all objective used by~\citet{ipad},
accumulated across stages:
\begin{equation}
  \mathcal{L}_{\mathrm{traj}}
  \;=\; \sum_{s=1}^{S} \gamma^{\,S-s}
    \min_{r=1,\ldots,R}\;\frac{1}{J}\sum_{j=1}^{J}
    \big\lVert \boldsymbol{\tau}^{(s)}_{r,j}
      - \boldsymbol{\tau}^{\mathrm{GT}}_{j} \big\rVert_{1},
  \label{eq:traj}
\end{equation}
where $\boldsymbol{\tau}^{\mathrm{GT}}$ denotes the logged ground-truth
trajectory and $\gamma$
discounts earlier stages. The minimum selects the closest proposal at each stage
for trajectory regression. Auxiliary heads supervise the expert's final motion states with category and
position losses ($\mathcal{L}_{\mathrm{agent}}^{\mathrm{aux}}$) and its final
meta-action states with a maneuver classification loss
($\mathcal{L}_{\mathrm{meta}}^{\mathrm{aux}}$).

\paragraph{Scene-grounded cross-proposal scoring.}
Dedicated score tokens incorporate scene and latent-state context when
evaluating trajectory proposals. Direct score-to-action attention is restricted to the
corresponding proposal, while attention among score tokens enables information
exchange across candidates (Fig.~\ref{fig:action_expert}(b)). An MLP maps the final
score states to scalar scores, supervised by binary cross-entropy
$\mathcal{L}_{\mathrm{score}}$ against simulated trajectory scores~\cite{navsim}.
At inference, the highest-scoring proposal is selected.

\begin{table*}[t]
\centering
\caption{\textbf{NAVSIM v1 \emph{navtest} results (PDMS).}
$n$C denotes $n$ cameras, and +L indicates LiDAR input. The FSDrive camera count follows DriveDreamer-Policy~\cite{drivedreamerpolicy}.
$^\dagger$ marks VLA models trained with reinforcement learning. GeoVLDrive uses supervised training.
Baseline scores come from the cited papers. We use the NAVSIM reimplementation of UniAD~\cite{navsim},
the camera--LiDAR VADv2-V$_{8192}$ from DiffusionDrive~\cite{diffusiondrive}, Hydra-MDP V$_{8192}$-W-EP,
WorldDrive trained on the full \emph{navtrain} split, and the multiple-anchor DriveVLA-W0 variant
with unspecified backbone (--). We exclude variants using an oracle to select sampled trajectories.
Bold marks the highest displayed values, including ties. Supplementary Table~\ref{tab:app_navsim_metrics} defines the metrics.}
\label{tab:navsim_v1}
\normalsize
\setlength{\tabcolsep}{6.0pt}
\begin{tabular}{@{}lcccccccc@{}}
\toprule
Method & \shortstack{Backbone} & Sensors & NC$\uparrow$ & DAC$\uparrow$ & TTC$\uparrow$
 & Comf.$\uparrow$ & EP$\uparrow$ & PDMS$\uparrow$ \\
\midrule
VADv2-V$_{8192}$~\cite{vadv2} & \multirow{5}{*}{E2E} & 3C+L & 97.2 & 89.1 & 91.6 & \textbf{100.0} & 76.0 & 80.9 \\
UniAD~\cite{uniad} &  & 8C & 97.8 & 91.9 & 92.9 & \textbf{100.0} & 78.8 & 83.4 \\
Hydra-MDP~\cite{hydramdp} &  & 3C+L & 98.3 & 96.0 & 94.6 & \textbf{100.0} & 78.7 & 86.5 \\
DiffusionDrive~\cite{diffusiondrive} &  & 3C+L & 98.2 & 96.2 & 94.7 & \textbf{100.0} & 82.2 & 88.1 \\
iPad~\cite{ipad} &  & 4C & 98.6 & 98.3 & 94.9 & \textbf{100.0} & 88.0 & 91.7 \\
\midrule
FSDrive~\cite{fsdrive} & \multirow{6}{*}{World model} & 3C & 98.2 & 93.8 & 93.3 & 99.9 & 80.1 & 85.1 \\
PWM~\cite{pwm} &  & 1C & 98.6 & 95.9 & 95.4 & \textbf{100.0} & 81.8 & 88.1 \\
WoTE~\cite{wote} &  & 3C+L & 98.5 & 96.8 & 94.9 & 99.9 & 81.9 & 88.3 \\
WorldDrive~\cite{worlddrive} &  & 1C & 98.4 & 96.8 & 95.2 & \textbf{100.0} & 83.3 & 89.0 \\
PerceptDrive~\cite{perceptdrive} &  & 1C & \textbf{99.2} & 99.0 & \textbf{98.0} & \textbf{100.0} & 82.9 & 90.4 \\
GeoWorldAD~\cite{geoworldad} &  & 1C & 99.0 & 97.8 & 95.8 & 99.9 & 85.9 & 91.0 \\
\midrule
DriveVLA-W0~\cite{drivevlaw0} & -- & 1C & 98.7 & \textbf{99.1} & 95.3 & 99.3 & 83.3 & 90.2 \\
LaST-VLA~\cite{lastvla} & InternVL3-8B & 1C & 98.7 & 95.4 & 95.6 & \textbf{100.0} & 80.5 & 87.3 \\
LaST-VLA$^\dagger$~\cite{lastvla} & InternVL3-2B & 1C & 98.6 & 97.7 & 95.6 & \textbf{100.0} & 86.5 & 91.1 \\
LaST-VLA$^\dagger$~\cite{lastvla} & InternVL3-8B & 1C & 98.7 & 97.9 & 95.6 & \textbf{100.0} & 86.8 & 91.3 \\
SGDrive (SFT)~\cite{sgdrive} & InternVL3-2B & 1C & 98.6 & 95.1 & 95.4 & \textbf{100.0} & 81.2 & 87.4 \\
VGGDrive~\cite{vggdrive} & Qwen2.5-VL-7B & 3C & 98.6 & 96.3 & 95.6 & \textbf{100.0} & 82.9 & 88.8 \\
CoWorld-VLA~\cite{coworldvla} & Qwen3-VL-2B & 1C & 99.1 & 97.0 & 96.5 & \textbf{100.0} & 84.0 & 90.0 \\
DriveWorld-VLA~\cite{driveworldvla} & InternVL3-2B & 3C & 99.1 & 98.2 & 96.1 & \textbf{100.0} & 85.9 & 91.3 \\
\midrule
\ourrow \textbf{GeoVLDrive} & Qwen3-VL-2B & 1C & 98.7 & 98.0 & 95.3 & \textbf{100.0} & \textbf{88.1} & \textbf{91.8} \\
\bottomrule
\end{tabular}
\end{table*}

\subsection{Training Objectives}
\label{sec:loss}

We combine latent-chain and textual CoT supervision with trajectory regression
and proposal scoring. Auxiliary losses supervise the action expert's refined
motion and meta-action states. The joint action-training objective is
\begin{equation}
  \mathcal{L}
  = \mathcal{L}_{\mathrm{traj}}
  + \sum_{q\in\mathcal{T}}\lambda_q\mathcal{L}_q
  + \sum_{q\in\{\mathrm{agent},\mathrm{meta}\}}
    \lambda_q^{\mathrm{aux}}\mathcal{L}_q^{\mathrm{aux}},
  \label{eq:total}
\end{equation}
where $\mathcal{T}=\{\mathrm{world},\mathrm{agent},\mathrm{meta},\mathrm{cot},\mathrm{score}\}$,
and the $\lambda$ coefficients weight the loss terms.
We use autoregressive cross-entropy on annotated textual reasoning for
$\mathcal{L}_{\mathrm{cot}}$. Prediction heads supervise the latent states
for geometry, motion, and maneuvers.

\section{Experiments}
\label{sec:exp}

\subsection{Implementation Details}
\label{sec:implementation}
\label{sec:data}

\paragraph{Datasets.}
We use LingoQA~\cite{lingoqa} and NAVSIM~3DQA for visual question answering.\footnote{
This work uses LingoQA and NAVSIM solely for academic research purposes.
The authors confirm that all uses of these datasets and benchmarks are strictly
limited to non-commercial academic research and do not involve any commercial
applications.}
NAVSIM-WM-CoT (Sec.~\ref{sec:app_chain} of the supplementary material) provides chain-of-thought
annotations and supervision for future geometry and key-agent motion. For planning, we train on \emph{navtrain} and
evaluate on the $12{,}146$ scenes in \emph{navtest}~\cite{navsim,navsimv2}.
Inference uses front-camera images and ego state without LiDAR.

\noindent\emph{NAVSIM 3DQA.}
We derive NAVSIM~3DQA from NAVSIM annotations to train and evaluate the
backbone on object geometry and motion. We will publicly release NAVSIM~3DQA
when we submit the camera-ready version.
Each sample comprises four consecutive front-camera frames at time offsets
$[-1.5,-1.0,-0.5,0]$\,s relative to the current frame, ordered from oldest to
newest, and a question about an object identified by a normalized $2$D
bounding box in the current frame. Questions cover four
task types: coarse object category, ground-plane position in meters, vehicle
dimensions, and motion behavior. The dataset contains approximately $1.28$M QA
pairs from $95$K anchor frames, including $1.12$M training pairs and $165$K
validation pairs. The training and validation sets follow NAVSIM's official
partition. Dataset construction is detailed in Sec.~\ref{sec:data_app} of the supplementary material.

\begin{table*}[t]
\centering
\caption{\textbf{NAVSIM v2 results (EPDMS).}
$^\dagger$ marks reinforcement learning.
Scores for Ego Status, TransFuser, and Hydra-MDP++ follow DriveVLA-W0~\cite{drivevlaw0},
and DiffusionDrive scores are from GeoWorldAD~\cite{geoworldad}.
Other baseline scores are taken from their cited papers.
$^\ddagger$ARTEMIS reports EC and labels a separate comfort measure C, so its HC entry is unspecified.
Bold marks the highest displayed values, including ties.}
\label{tab:navsim_v2}
\normalsize
\setlength{\tabcolsep}{5.4pt}
\begin{tabular}{@{}lcccccccccc@{}}
\toprule
Method & NC$\uparrow$ & DAC$\uparrow$ & DDC$\uparrow$ & TLC$\uparrow$
 & EP$\uparrow$ & TTC$\uparrow$ & LK$\uparrow$ & HC$\uparrow$ & EC$\uparrow$
 & EPDMS$\uparrow$ \\
\midrule
Ego Status~\cite{navsimv2} & 93.1 & 77.9 & 92.7 & 99.6 & 86.0 & 91.5 & 89.4 & 98.3 & 85.4 & 64.0 \\
TransFuser~\cite{transfuser} & 96.9 & 89.9 & 97.8 & 99.7 & 87.1 & 95.4 & 92.7 & 98.3 & 87.2 & 76.7 \\
Hydra-MDP++~\cite{hydramdppp} & 97.2 & 97.5 & 99.4 & 99.6 & 83.1 & 96.5 & 94.4 & 98.2 & 70.9 & 81.4 \\
ARTEMIS$^\ddagger$~\cite{artemis} & 98.3 & 95.1 & 98.6 & 99.8 & 81.5 & 97.4 & 96.5 & -- & \textbf{98.3} & 83.1 \\
DiffusionDrive~\cite{diffusiondrive} & 98.2 & 95.9 & 99.4 & 99.8 & 87.5 & 97.3 & 96.8 & 98.3 & 87.7 & 84.5 \\
PerceptDrive~\cite{perceptdrive} & 98.6 & 98.8 & 99.5 & \textbf{99.9} & 87.5 & 97.9 & 96.9 & \textbf{98.5} & 86.3 & 90.2 \\
GeoWorldAD~\cite{geoworldad} & \textbf{99.0} & 97.8 & \textbf{99.6} & 99.7 & 89.1 & \textbf{98.6} & \textbf{97.6} & 98.0 & 82.2 & 90.4 \\
\midrule
DriveVLA-W0~\cite{drivevlaw0} & 98.5 & \textbf{99.1} & 98.0 & 99.7 & 86.4 & 98.1 & 93.2 & 97.9 & 58.9 & 86.1 \\
DriveWorld-VLA~\cite{driveworldvla} & 98.6 & \textbf{99.1} & \textbf{99.6} & 99.8 & 87.4 & 97.9 & 97.0 & 97.8 & 78.6 & 86.8 \\
LaST-VLA-8B$^\dagger$~\cite{lastvla} & 98.7 & 97.9 & 99.2 & 99.7 & 90.3 & 98.2 & 96.6 & 98.3 & 86.3 & 87.1 \\
\midrule
\ourrow \textbf{GeoVLDrive} & 98.7 & 98.0 & 99.5 & 99.6 & \textbf{91.4} & 98.2 & \textbf{97.6} & 97.9 & 82.4 & \textbf{90.9} \\
\bottomrule
\end{tabular}
\end{table*}

\paragraph{Metrics.}
We report the Predictive Driver Model Score (PDMS) on NAVSIM~v1 and the
Extended PDMS (EPDMS) on NAVSIM~v2 for planning evaluation. Both aggregate
safety, comfort, and progress; v2 includes additional evaluation criteria.
We evaluate NAVSIM~v2 metrics on the \emph{navtest} split using single-stage simulation.
NAVSIM~3DQA scores four tasks: coarse category and motion behavior by exact
match of parsed semantic labels, ground-plane position by a distance-dependent error tolerance, and
vehicle dimensions by a fixed $0.5$\,m tolerance. Overall is the
equal-weight mean of the four task scores. The displayed Position and Dimensions
columns use separate diagnostic criteria, as detailed in
Sec.~\ref{sec:app_qa_eval} of the supplementary material.
On LingoQA we report Lingo-Judge accuracy.

\paragraph{Training details.}
We use Qwen3-VL-2B-Instruct~\cite{qwen3vl} as the vision-language backbone and
StreamVGGT~\cite{streamvggt} as the geometry encoder, which remains frozen
throughout training. Training proceeds in four phases. The first aligns
semantics and geometry on the question-answering mixture for $3$ epochs. The
second supervises future geometry and key-agent motion for $5$ epochs. The
third trains the supervised latent chain for $10$ epochs, and the fourth trains
the chain and the action expert jointly for $30$ epochs. Training uses
NVIDIA H20 GPUs, with phase-specific GPU counts and batch sizes listed in
supplementary Table~\ref{tab:train_recipe}. Sec.~\ref{sec:impl} of the supplementary material describes the latent
interface, loss definitions, and training schedule.

\begin{table}[t]
\centering
\caption{\textbf{LingoQA and NAVSIM 3DQA results (\%, higher is better).}
Qwen3-VL~\cite{qwen3vl} rows use the Instruct models. InternVL3.5~\cite{internvl35} and Qwen3-VL in the upper block have no task-specific fine-tuning.
$^\ast$ marks the fine-tuned baseline. We train the geometry-grounded GeoVLDrive backbone with the same data and schedule.
See supplementary Sec.~\ref{sec:app_qa_eval} for Overall and Tables~\ref{tab:3dqa_category_acc}--\ref{tab:3dqa_motion_acc} for per-class accuracies.
Position accuracy uses $L_2\!\leq\!1$\,m. Size accuracy requires all three errors to be within $0.5$\,m.}
\label{tab:3dqa}
\small
\setlength{\tabcolsep}{2.1pt}
\begin{tabular}{@{}lcccccc@{}}
\toprule
& \textbf{LingoQA} & \multicolumn{5}{c}{\textbf{NAVSIM 3DQA}} \\
\cmidrule(lr){2-2}\cmidrule(lr){3-7}
Method & Judge & Overall & Cat. & Mot. & Pos. & Dim. \\
\midrule
Qwen3-VL-2B & 44.0 & 34.2 & 80.3 & 38.5 & 6.1 & 10.1 \\
Qwen3-VL-8B & 54.8 & 54.4 & 82.3 & 50.7 & 9.4 & 68.7 \\
InternVL3.5-2B       & 53.8 & 29.9 & 66.9 & 27.3 & 5.0 & 8.4 \\
InternVL3.5-8B       & 51.4 & 45.1 & 67.0 & 33.5 & 7.4 & 62.5 \\
\midrule
Qwen3-VL-2B$^\ast$ & 67.0 & 84.6 & 97.9 & 85.8 & 66.8 & 87.8 \\
\ourrow \textbf{GeoVLDrive} & \textbf{68.6} & \textbf{89.2}
 & \textbf{98.0} & \textbf{91.0} & \textbf{76.7} & \textbf{89.1} \\
\bottomrule
\end{tabular}
\end{table}

\subsection{Evaluation Results}
\label{sec:results}

\paragraph{NAVSIM benchmark.}
Using a 2B VLM backbone, one front camera, and supervised training only,
GeoVLDrive achieves $91.8$ PDMS on NAVSIM~v1 and $90.9$ EPDMS on NAVSIM~v2
(Tables~\ref{tab:navsim_v1} and~\ref{tab:navsim_v2}). On NAVSIM~v1, it exceeds
the 2B-backbone VLAs SGDrive (SFT)~\cite{sgdrive}, CoWorld-VLA~\cite{coworldvla},
and DriveWorld-VLA~\cite{driveworldvla} by $4.4$, $1.8$, and $0.5$ PDMS, and the
RL-trained LaST-VLA-2B~\cite{lastvla} by $0.7$. Among proposal-based planners,
iPad~\cite{ipad} scores $91.7$ PDMS and GeoWorldAD~\cite{geoworldad} scores
$91.0$. Additional planning visualizations are provided in supplementary Sec.~\ref{sec:app_qual_cases}.

\paragraph{LingoQA and NAVSIM 3DQA.}
We compare the geometry-grounded backbone of GeoVLDrive with
Qwen3-VL-2B-Instruct fine-tuned on the same data and schedule
(Table~\ref{tab:3dqa}). GeoVLDrive improves the overall NAVSIM~3DQA score from
$84.6$\% to $89.2$\%, with gains in metric position and motion accuracy. On
LingoQA, it achieves $68.6$\% Lingo-Judge accuracy compared with $67.0$\% for
the matched fine-tuned baseline, showing a modest numerical increase in driving
question answering.

We evaluate future depth, key-agent motion, and meta action on the
NAVSIM-WM-CoT validation split (supplementary Tables~\ref{tab:intermediate_predictions} and~\ref{tab:future_depth_val}).

\begin{table}[t]
\centering
\caption{\textbf{Geometry injection and spatial alignment.}
Geo. adds StreamVGGT geometry injection. UVPos adds normalized $(u,v)$ positional encoding. The baseline is Qwen3-VL-2B-Instruct$^\ast$.
Metrics follow Table~\ref{tab:3dqa} (\%, higher is better).}
\label{tab:ablation_geo}
\small
\setlength{\tabcolsep}{2.25pt}
\begin{tabular}{@{}lcccccc@{}}
\toprule
& \textbf{LingoQA} & \multicolumn{5}{c}{\textbf{NAVSIM 3DQA}} \\
\cmidrule(lr){2-2}\cmidrule(lr){3-7}
Configuration & Judge & Overall & Cat. & Mot. & Pos. & Dim. \\
\midrule
Baseline$^\ast$ & 67.0 & 84.6 & 97.9 & 85.8 & 66.8 & 87.8 \\
+ Geo.            & 67.6 & 87.5 & 97.9 & 90.1 & 72.2 & 88.4 \\
\ourrow + Geo. + UVPos & \textbf{68.6} & \textbf{89.2}
 & \textbf{98.0} & \textbf{91.0} & \textbf{76.7} & \textbf{89.1} \\
\bottomrule
\end{tabular}
\end{table}

\begin{table}[t]
\centering
\caption{\textbf{Geometry-grounded latent CoT ablations on NAVSIM v1.}
  All configurations use the same action-expert architecture.
  (a)~World, Motion, Meta, and Geo. enable future geometry, key-agent motion, meta action, and geometry injection.
  (b)~We compare the directed mask in Fig.~\ref{fig:mask} with a variant that allows bidirectional attention between the same latent CoT groups.
  Directed repeats the full-model result from (a).
  Bold marks the highest values within each subtable, including ties.}
\label{tab:ablation_components}
\small
\setlength{\tabcolsep}{2.05pt}
\begin{tabular}{@{}cccccccccc@{}}
\toprule
\multicolumn{10}{@{}l}{\emph{(a) Latent groups and geometry injection}} \\
World & Motion & Meta & Geo. & NC & DAC & TTC & Comf. & EP & PDMS \\
\midrule
&            &            &            & 98.2 & 97.3 & 93.6 & 99.9 & 87.0 & 90.1 \\
\checkmark &            &            &            & 98.4 & 97.7 & 93.8 & \textbf{100.0} & 87.0 & 90.5 \\
\checkmark & \checkmark &            &            & 98.8 & 97.5 & 95.3 & 99.9 & 86.1 & 90.7 \\
\checkmark & \checkmark & \checkmark &            & \textbf{99.0} & 97.8 & \textbf{95.7} & 99.9 & 86.0 & 91.0 \\
\ourrow \checkmark & \checkmark & \checkmark & \checkmark
 & 98.7 & \textbf{98.0} & 95.3 & \textbf{100.0} & \textbf{88.1} & \textbf{91.8} \\
\midrule
\multicolumn{10}{@{}l}{\emph{(b) Attention configurations}} \\
\multicolumn{4}{@{}l}{Bidirectional} & 98.5 & 97.4 & 94.8 & 99.9 & 87.1 & 90.8 \\
\ourrow \multicolumn{4}{@{}l}{Directed (ours)}
 & \textbf{98.7} & \textbf{98.0} & \textbf{95.3} & \textbf{100.0} & \textbf{88.1} & \textbf{91.8} \\
\bottomrule
\end{tabular}
\end{table}

\begin{table}[t]
\centering
\caption{\textbf{Trajectory scoring on NAVSIM v1.} The MLP reads max-pooled action states, following iPad~\cite{ipad}.
For score tokens, Scene enables direct scene attention and Cross enables attention among score tokens across proposals.
Checkmarks indicate enabled paths, blanks indicate disabled paths, and -- means inapplicable. Higher scores are better.}
\label{tab:ablation_head}
\small
\setlength{\tabcolsep}{1.7pt}
\begin{tabular}{@{}lcccccccc@{}}
\toprule
Scorer & Scene & Cross & NC & DAC & TTC & Comf. & EP & PDMS \\
\midrule
MLP & -- & -- & 98.4 & 97.4 & 94.3 & 99.9 & 86.4 & 90.2 \\
Score tokens &    &    & 98.5 & 97.4 & 93.9 & 99.9 & 87.0 & 90.4 \\
Score tokens & \checkmark &    & 98.6 & 97.7 & 94.2 & \textbf{100.0} & 87.2 & 90.8 \\
\ourrow Score tokens & \checkmark & \checkmark
 & \textbf{98.7} & \textbf{98.0} & \textbf{95.3} & \textbf{100.0} & \textbf{88.1} & \textbf{91.8} \\
\bottomrule
\end{tabular}
\end{table}

\subsection{Ablation Study}
\label{sec:ablation}
\paragraph{Geometry injection and spatial alignment.}
Geometry injection improves the NAVSIM~3DQA score from $84.6$ to $87.5$, and
normalized $(u,v)$ alignment further raises it to $89.2$
(Table~\ref{tab:ablation_geo}). Position accuracy increases from $66.8$ to
$72.2$ and $76.7$, respectively. LingoQA accuracy also rises from $67.0$ to
$67.6$ and $68.6$.

\paragraph{Geometry-grounded latent CoT.}
With the same action-expert architecture, adding World, Motion, and Meta to the
pre-aligned backbone raises PDMS from $90.1$ to $90.5$, $90.7$, and $91.0$,
respectively (Table~\ref{tab:ablation_components}(a)). NC and TTC rise from
$98.2$ to $99.0$ and $93.6$ to $95.7$, respectively.
With all three latent groups enabled, adding geometry injection raises PDMS
from $91.0$ to $91.8$. The largest change is in ego progress ($+2.1$ points),
with changes of $+0.2$ in DAC, $-0.3$ in NC, and $-0.4$ in TTC.
Directed cross-group attention yields $91.8$ PDMS,
compared with $90.8$ for bidirectional attention (Table~\ref{tab:ablation_components}(b)).

\paragraph{Proposal scoring.}
With all other settings fixed, Table~\ref{tab:ablation_head} compares an iPad-style
MLP readout of max-pooled action states~\cite{ipad} with three score-token variants.
PDMS rises from $90.2$ (MLP) to $90.4$ (action-only tokens), $90.8$ (scene-aware
tokens), and $91.8$ (cross-proposal attention).
Cross-proposal attention improves NC, DAC, TTC, and EP over scene-aware scoring
alone. These results support incorporating interactions among candidates when
evaluating trajectory proposals.

\FloatBarrier
\section{Conclusion}

GeoVLDrive combines multi-layer geometry injection with latent-state supervision
for future scene geometry, key-agent motion, and meta action. The action expert uses these
states and multi-depth scene features to generate and score trajectory
proposals. The latent CoT and geometry injection
raise NAVSIM~v1 PDMS from $90.1$ to $91.8$, and the full model
reaches $90.9$ EPDMS on NAVSIM~v2 with one front camera and supervised training
only. Scene-grounded, interacting score tokens add $1.6$ PDMS over an MLP readout.
Geometry injection improves
box-conditioned metric localization and motion recognition on~NAVSIM~3DQA.

\paragraph{Limitations and future work.}
GeoVLDrive uses a single front camera and models at most one key agent per
scene. Future geometry is supervised over $2$\,s, while key-agent motion and
planning extend to $4$\,s. Training targets follow the recorded future. We evaluate with
NAVSIM's non-reactive pseudo-simulation. Future work includes surround-view input, multi-agent interactions, longer
geometry horizons, multimodal futures, and closed-loop evaluation.

\section*{LLM Usage Statement}
The authors used OpenAI Codex and Anthropic Claude to polish the manuscript.
OpenAI Codex also assisted with cleaning the training data.
We used OpenAI GPT to check and correct textual CoT annotations against the
structured scene annotations. Human annotators reviewed a sample of the revised texts.

{\small
\bibliographystyle{ieeenat_fullname}
\bibliography{references}
}

\arxivSupplement
\section{NAVSIM 3DQA}
\label{sec:appendix}
\label{sec:data_app}
\label{sec:app_3dqa}

\subsection{Dataset Construction}
\label{sec:app_3dqa_construction}

\begin{figure}[p]
\centering
\includegraphics[width=\linewidth,height=.86\textheight,keepaspectratio]{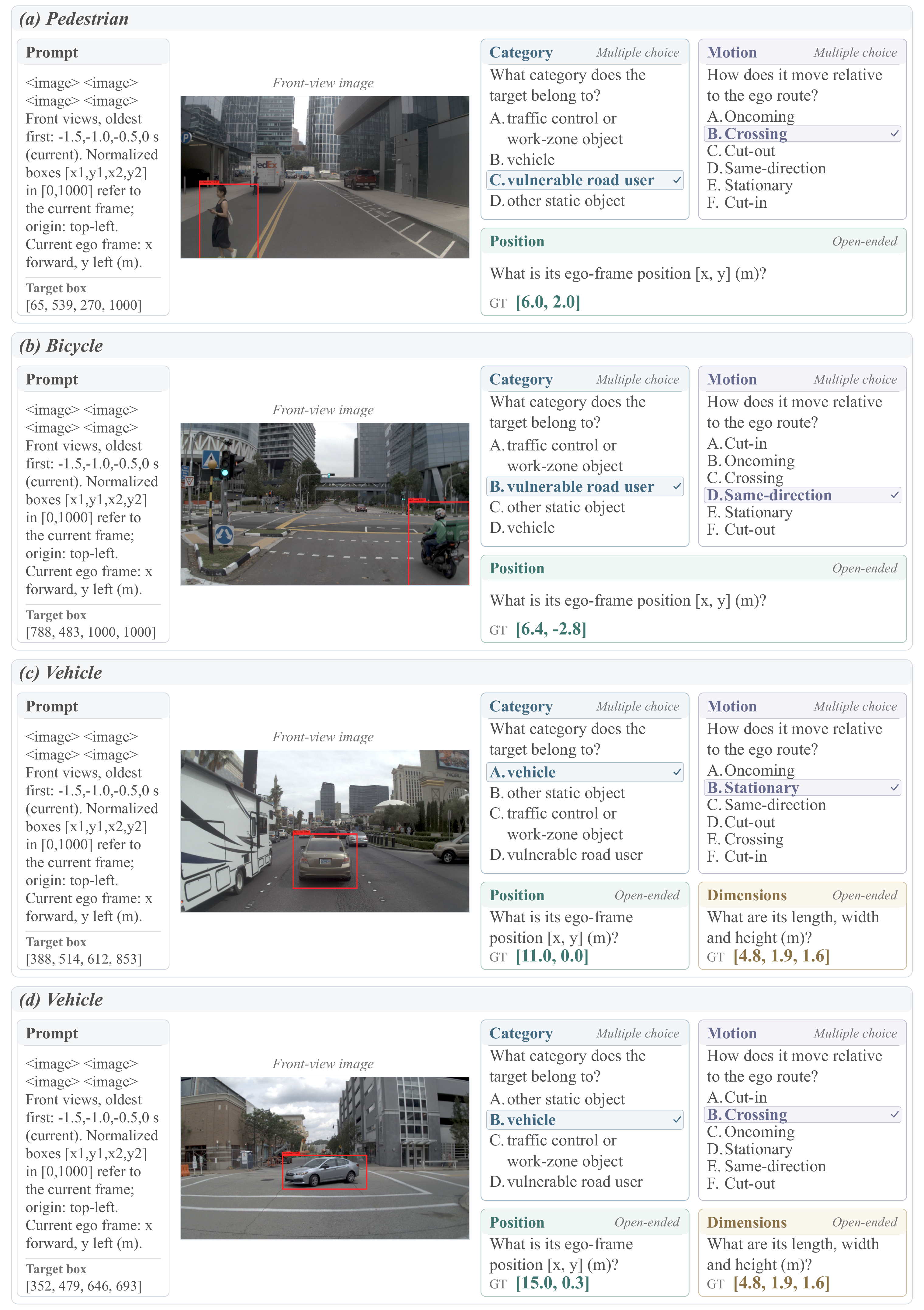}
\caption{\textbf{NAVSIM 3DQA examples.} Each row shows the shared prompt,
target box, current front view, and questions. Only the current frame is shown;
the three history frames are omitted. Boxes and ego coordinates refer to the
current frame. Colors distinguish question types. Options retain their original
order; correct answers are outlined and checked. Numerical answers are labeled
GT (ground truth). The prompt and questions are abbreviated for display.}

\label{fig:app_3dqa_examples}
\end{figure}

\paragraph{Data sources and object grounding.}
We construct questions from NAVSIM front-camera observations, camera
calibration, annotated $3$D boxes, instance tracks, and HD maps. Target objects
are selected independently of the latent-chain annotations. Each cuboid is
projected by clipping its twelve edges at a $0.1$\,m near plane, sampling nine
points per retained edge, and applying camera distortion and intrinsics.
The projected convex hull is clipped to the image, and its axis-aligned box
is normalized to $[0,1000]^2$ to identify the target in the text prompt.
Model inputs contain unannotated images; the box overlays in
Fig.~\ref{fig:app_3dqa_examples} are used only for visualization.

\paragraph{Visibility filtering.}
Candidate objects must belong to a supported class, have positive box
dimensions, and lie ahead of the ego vehicle within a planar distance
$d=\sqrt{c_x^2+c_y^2}\le80$\,m, where $(c_x,c_y)$ is the box center in the
current ego frame. Let $P$ and $P'$ be the projected hull before and after
image clipping, with widths $w$ and $w'$. Horizontal truncation
$T_h=1-w'/w$ and area truncation $T_a=1-|P'|/|P|$ must each be at most $0.5$;
$|P|$ denotes polygon area. We approximate occlusion using overlaps with
objects more than $0.05$\,m closer in camera depth:
\begin{equation}
O_i=\frac{\bigl|\bigcup_{j:\,z_j<z_i-0.05}(P'_i\cap P'_j)\bigr|}{|P'_i|}\le0.5,
\end{equation}
where $z_i$ is the camera-depth coordinate of object $i$.
At a reference resolution of $280\times518$ (height $\times$ width), the
minimum box short sides are $10$, $12$, and $16$ pixels and the minimum areas
are $160$, $250$, and $400$ pixels squared for $d\le40$\,m,
$40<d\le60$\,m, and $60<d\le80$\,m, respectively. Both thresholds must be met.

\paragraph{Motion annotation.}
Vehicle, pedestrian, and bicycle tracks are transformed to the ego frame at
the anchor time. For an object center in homogeneous coordinates,
\begin{equation}
\bar{\mathbf{p}}_i^{(0)}(t)
=\mathbf{T}_{\mathrm{ego}(0)\leftarrow\mathrm{global}}\,
\mathbf{T}_{\mathrm{global}\leftarrow\mathrm{ego}(t)}\,
\bar{\mathbf{p}}_i^{(t)}(t),
\end{equation}
where $\mathbf{T}$ denotes a rigid coordinate transformation. This removes
ego motion, so \emph{stationary} refers to motion relative to the ground.
Each motion label requires at least three valid track states, including the
anchor, spanning at least one second. We fit a linear trajectory to three
states and a quadratic trajectory to four or more, using its derivative at
the anchor to estimate velocity. Fits with residuals above $0.8$\,m or
direction discrepancies above $40^\circ$ relative to the annotated velocity
are marked ambiguous; insufficient evidence is marked unavailable. These
observations remain eligible for other question types.

Motion classification combines lane overlap, centerline distance, and local
road direction (Table~\ref{tab:app_motion}). Cut-in and cut-out labels require
confirmation from annotations up to $2$\,s into the future; other boundary
cases, except crossing, may also use this evidence. The logged future ego
trajectory is excluded from label construction. Model observations are
restricted to the four front-camera frames at $[-1.5,-1.0,-0.5,0]$\,s.

\begin{table}[!htbp]
\supplementtableformat
\setlength{\tabcolsep}{4pt}
\caption{\textbf{Motion categories and annotation criteria.} The angle
$\theta$ measures the difference between the object's direction of travel
and the local road tangent. Cut-in and cut-out additionally require the
direction of travel to be within $55^\circ$ of the ego forward direction.}

\label{tab:app_motion}
\begin{tabular}{P{.17\linewidth}P{.76\linewidth}}
\toprule
Class & Annotation criteria \\
\midrule
\emph{stationary} & Low ground speed and small recent displacement; selected
low-speed cases additionally use future displacement and terminal speed for
confirmation. \\
\addlinespace[2pt]
\emph{same direction} & Motion aligned with the local road tangent
($\theta\le40^\circ$). \\
\addlinespace[2pt]
\emph{oncoming} & Motion opposing the local road tangent
($\theta\ge140^\circ$). \\
\addlinespace[2pt]
\emph{crossing} & Lateral motion at or through the ego road corridor
($55^\circ\le\theta\le125^\circ$). Pedestrian annotation also uses road-surface
and crosswalk occupancy. \\
\addlinespace[2pt]
\emph{cut-in} & Vehicle or bicycle entering the ego lane over the observed
history, confirmed by future lane occupancy and motion direction. \\
\addlinespace[2pt]
\emph{cut-out} & Vehicle or bicycle leaving the ego lane over the observed
history, with the same future-confirmation criteria. \\

\bottomrule
\end{tabular}
\end{table}

\paragraph{Event grouping and object selection.}
Cut-in and cut-out observations are grouped by scene, track, and label, with
temporal gaps above $1.25$\,s defining separate events. Each event contributes
one motion question, selected by annotation confidence, visibility, transition
stage, and salience. Other frames of the same event remain eligible for
other question types.
Each frame retains at most five objects, prioritizing up to two event anchors,
the two most salient candidates, and the nearest eligible object. Remaining
slots promote category and spatial coverage, prioritizing vulnerable road
users and traffic-control objects over vehicles, followed by fine-grained
category diversity. Spatial bins partition longitudinal position $c_x$ at
$[20,40,60]$\,m and lateral position $c_y$ at $[-6,-2,2,6]$\,m.
The salience score is
\begin{equation}
S_i=\log(1+A_i)-0.0125\,d_i-1.5\,O_i-T_{a,i}-0.5\,T_{h,i},
\end{equation}
where $A_i$ is the box area at the reference resolution. Ties are resolved
deterministically using a stable hash. Coverage-based selection excludes
candidates dominated by a same-category object that is at least four times
larger in projected area, closer, and no more occluded or truncated.

\paragraph{Question generation.}
We derive the four tasks in Table~\ref{tab:app_3dqa_tasks} from structured
annotations. Prompts specify coordinates and units. A stable hash determines
the order of multiple-choice options.
The coarse categories are vehicle, vulnerable road user (pedestrian or
bicycle), traffic-control or work-zone object (cone, barrier, or
construction-zone sign), and other static object. Category questions identify
the target by its box; other tasks also provide its class. Every selected
object contributes category and position questions. Vehicles additionally
contribute dimension questions, and objects with valid motion labels
contribute motion questions, yielding at most twenty questions per frame.
Numerical answers are rounded to one decimal place. Evaluation follows
Appendix~\ref{sec:app_qa_eval}.

\begin{table}[!htbp]
\supplementtableformat
\setlength{\tabcolsep}{4pt}
\caption{\textbf{NAVSIM 3DQA tasks and outputs.} Positions use the current
ego frame; all numerical outputs are in meters. The tasks exclude vertical
position and object yaw. Accuracy criteria are defined in
Appendix~\ref{sec:app_qa_eval}.}

\label{tab:app_3dqa_tasks}
\begin{tabular}{P{.24\linewidth}P{.27\linewidth}P{.18\linewidth}P{.23\linewidth}}
\toprule
Task & Eligible objects & Output & Evaluation \\
\midrule
Coarse category & All selected objects & Category label & Accuracy \\
\addlinespace[2pt]
Ground-plane position & All selected objects & $[x,y]$ & Axis / $L_2$ accuracy \\
\addlinespace[2pt]
Vehicle dimensions & Vehicles & $[\ell,w,h]$ & Component / joint accuracy \\
\addlinespace[2pt]
Motion behavior & Vehicles, pedestrians, and bicycles with valid labels
& Motion label & Accuracy \\

\bottomrule
\end{tabular}
\end{table}

\subsection{Dataset Statistics}
\label{sec:app_3dqa_stats}

NAVSIM 3DQA contains $1{,}281{,}855$ QA pairs: $1{,}117{,}204$ for training
and $164{,}651$ for validation. Of the $85{,}070$ training and $18{,}179$
validation source anchors, $80{,}161$ and $14{,}467$ yield eligible questions.
The two splits contain $333{,}920$ and $48{,}958$ object observations,
covering $89{,}533$ and $12{,}792$ distinct tracks, respectively. An object
observation denotes one object at one anchor frame. The source split is
preserved, with no overlap in scene, frame, track, or log-segment identifiers.

\begin{figure}[!htbp]
\centering
\includegraphics[width=\linewidth]{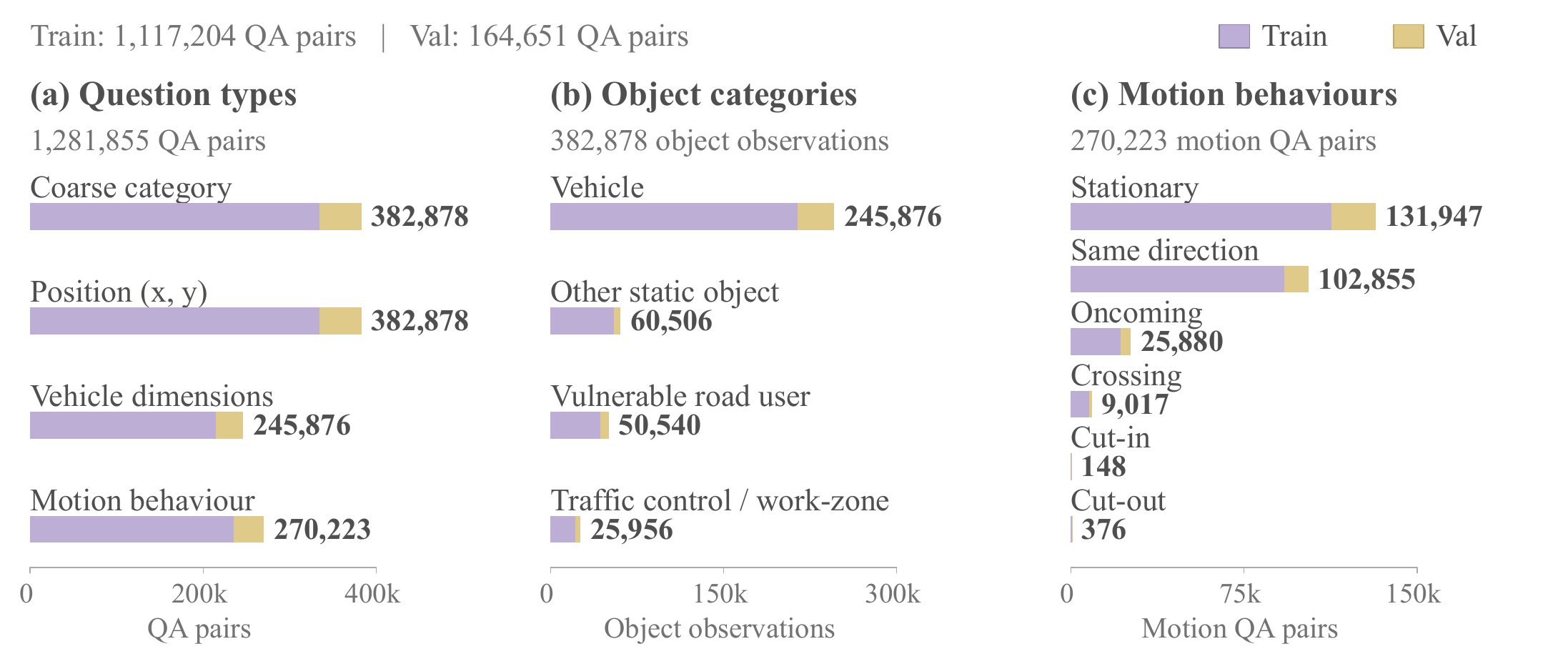}
\caption{\textbf{NAVSIM 3DQA composition.} Distributions of question types,
object categories, and motion behaviors. Stacked bars show training (lavender)
and validation (pale yellow) counts; labels report their sum. Object counts
refer to object--anchor observations. All axes use a linear scale.}

\label{fig:app_3dqa_statistics}
\end{figure}

Figure~\ref{fig:app_3dqa_statistics} summarizes the dataset composition.
Stationary and same-direction labels account for $86.89$\% of motion questions;
cut-in and cut-out contribute $524$ ($0.19$\%). Ambiguous observations represent
$2.6$\% of labeled or ambiguous candidates, and unavailable observations
represent $8.5$\% of all dynamic candidates. Automated audits verified that
each frame retains its two most salient and nearest eligible objects; the
largest projected box is retained in over $99.5$\% of frames. A stratified
visual audit of $40$ validation frames found no visibility, projection, or
scene-consistency issues.

\subsection{Evaluation Protocol}
\label{sec:app_qa_eval}

\paragraph{Task scores.}
Responses are parsed deterministically, with parsing failures counted as
incorrect. Category and motion accuracy are computed after mapping responses
to semantic labels. Tables~\ref{tab:3dqa_category_acc} and~\ref{tab:3dqa_motion_acc}
report validation accuracy by object category and motion class, respectively.
Bold marks the highest accuracy in each column, including ties.
A position answer is correct when both coordinate errors
are within the distance-dependent tolerance: $0.5$\,m for $d\le30$\,m,
$1$\,m for $30<d\le60$\,m, and $2$\,m for $d>60$\,m. Dimension accuracy
averages the correctness of length, width, and height, each with an absolute
error tolerance of $0.5$\,m. Each task is averaged over its eligible questions.
For task accuracies $A_{\mathrm{cat}}$, $A_{\mathrm{motion}}$,
$A_{\mathrm{pos}}^{\mathrm{axis}}$, and
$A_{\mathrm{dim}}^{\mathrm{component}}$ in $[0,1]$, the aggregate is
\begin{equation}
\mathrm{Overall}
=\frac{100}{4}\left(A_{\mathrm{cat}}+A_{\mathrm{motion}}
+A_{\mathrm{pos}}^{\mathrm{axis}}+A_{\mathrm{dim}}^{\mathrm{component}}\right).
\label{eq:app_qa_overall}
\end{equation}

\paragraph{Reported position and dimension diagnostics.}
In the NAVSIM 3DQA results table in the main paper, Position reports the fraction of answers with
ground-plane error $\|\hat{\mathbf{p}}-\mathbf{p}\|_2\le1$\,m, while
Dimensions requires all three component errors to be at most $0.5$\,m.
Overall uses the adaptive per-axis position accuracy and per-component
dimension accuracy defined above, so it is not the arithmetic mean of the
four displayed columns.

\subsection{Comparison with SURDS}
\label{sec:app_3dqa_surds}

NAVSIM 3DQA complements SURDS~\citep{surds} by emphasizing metric geometry
and temporal motion. SURDS evaluates orientation, depth ranges, pixel
localization, and pairwise spatial relations from individual nuScenes images,
using categorical correctness and a centerness-based localization score.
NAVSIM 3DQA uses four consecutive front views and a normalized $2$D box to
specify the queried object. It adds numerical estimation of ground-plane
position and vehicle dimensions in meters, together with temporal motion
classification.

The geometric diagnostics assess physical location and scale. Position
accuracy at $L_2\le1$\,m jointly constrains longitudinal and lateral errors in
the ego frame; joint dimension accuracy requires length, width, and height
all to be within $0.5$\,m of their targets. These criteria distinguish metric
accuracy from correct depth-range or relative-order predictions, which can
remain correct despite substantial coordinate errors. Estimating physical
dimensions further probes object scale, which image-plane localization alone
does not measure. Both tasks require numerical answers without candidate
values supplied in the question.

Motion classification extends the evaluation to temporal observations, with
labels derived from ego-motion-compensated tracks and local road geometry.
Categories such as stationary, crossing, and cut-in are therefore defined by
object motion relative to the ground and the road rather than by apparent
image motion. Together, the tasks provide complementary diagnostics of metric
spatial grounding and temporal motion understanding relevant to planning,
alongside SURDS's orientation and multi-object relation tests.

\section{Implementation and Training Details}
\label{sec:impl}

\subsection{Architecture and Latent Interface}
\label{sec:app_architecture}

\paragraph{Backbone configuration.}
StreamVGGT features from layers $[4,11,17,23]$ are injected into Qwen3-VL
layers $[6,13,20,26]$, respectively. The VLM and geometry streams use image
resolutions of $416\times672$ and $280\times518$ (height $\times$ width),
with a $20\times37$ patch grid in the geometry stream. Features across the
selected geometry layers differ in representation depth while sharing the
same spatial resolution.

\paragraph{Latent token interface.}
The observed visual and language prefix is followed by an ego-state token,
future-geometry tokens, key-agent motion tokens, and meta-action tokens.
Group-specific embeddings replace the reserved-token embeddings at these positions
in the VLM input sequence. With $\delta=0.5$\,s, future geometry uses $K=4$
horizons and $32$ tokens per horizon ($128$ in total). Agent motion uses ten
tokens for category, current position, and eight current-relative future
offsets. Two tokens encode longitudinal and lateral meta actions.
Future-geometry and future-motion tokens receive sinusoidal encodings of
their prediction times, with attention governed by the latent CoT mask in Sec.~\ref{sec:method} of the main paper.
Textual CoT targets follow the latent block during training. At inference,
planning uses the latent states without autoregressive text generation.

\paragraph{Coordinate frames.}
Agent positions and ego trajectories are expressed in meters in the ego
frame at the observation time, with $x$ forward and $y$ left. Future agent
annotations are transformed to this fixed frame before computing the
current-relative position offsets
$\Delta\mathbf{p}_j=\mathbf{p}_j-\mathbf{p}_0$, where $\mathbf{p}_0$ is the
agent's current position.

\paragraph{Action expert.}
The expert comprises four refinement stages, each containing four
independently parameterized attention blocks conditioned on features from
different depths. The corresponding Qwen3-VL readout indices are $[8,16,22,-1]$
(zero-based), with $-1$ selecting the final hidden state.
It maintains $R=64$ proposals, each represented by $J=8$
action tokens and one score token. Reasoning, action, and score states are
propagated between blocks, while scene features are supplied at each depth.
Action queries incorporate ego-state and meta-action context. VLM features
and latent states are detached before the expert's trainable projections,
blocking expert gradients from updating the VLM while allowing refinement
within the expert.

\subsection{Supervision and Loss Definitions}
\label{sec:app_losses}

\paragraph{Future geometry.}
We project recorded future LiDAR points into the front-camera image plane to
obtain sparse depth targets. Let $V_k$ denote the valid target pixels at
horizon $k$. For a depth map $\mathbf{d}$, define
\[
\begin{aligned}
s_k(\mathbf{d})
&=\max\!\left(\frac{\sum_{i\in V_k}|d_i|}{|V_k|+\epsilon},\epsilon\right),
\qquad \epsilon=10^{-8},\\
\tilde{\mathbf{z}}_k
&=\frac{\mathbf{z}_k}{s_k(\mathbf{z}_k)},\qquad
\tilde{\hat{\mathbf{z}}}_k
=\frac{\hat{\mathbf{z}}_k}{s_k(\hat{\mathbf{z}}_k)}.
\end{aligned}
\]
Target and predicted maps are thus normalized independently for each sample
and horizon. For the per-pixel error
$e_{k,i}=|\tilde{\hat z}_{k,i}-\tilde z_{k,i}|$, the regression and confidence
terms of the future-geometry objective are
\begin{equation}
\mathcal{L}_{\mathrm{depth}}^{(k)}
=\frac{1}{|V_k|}\sum_{i\in V_k}e_{k,i},\qquad
\mathcal{L}_{\mathrm{conf}}^{(k)}
=\frac{1}{|V_k|}\sum_{i\in V_k}
\left(c_{k,i}e_{k,i}-\alpha\log c_{k,i}\right),
\label{eq:app_world_depth}
\end{equation}
where $c_{k,i}>0$ is predicted confidence and $\alpha=0.2$ controls its
regularization. Gradient consistency averages four scales with subsampling
strides $[1,2,4,8]$. At each scale, we sum the absolute horizontal and vertical
finite differences of the signed normalized depth residual, retaining only
pairs with valid depths at both endpoints and clipping each difference at $100$.
The sum is normalized by the total number of valid pixels across the batch
at that scale. Depth and confidence losses average over all valid pixels
pooled across the batch at each horizon, after per-sample normalization.
The gradient term is not confidence-weighted.

\paragraph{Key-agent motion.}
Category supervision uses cross-entropy, including the \emph{none} class.
Let $\operatorname{SL1}$ denote Smooth L1 with mean reduction. Position
supervision averages the current-position, offset, and reconstructed-position
losses:
\begin{equation}
\mathcal{L}_{\mathrm{SmoothL1}}
=\frac{1}{3}\left[
\operatorname{SL1}(\hat{\mathbf{p}}_0,\mathbf{p}_0)
+\operatorname{SL1}(\widehat{\Delta\mathbf{p}}_{1:J},\Delta\mathbf{p}_{1:J})
+\operatorname{SL1}(\hat{\mathbf{p}}_{0:J},\mathbf{p}_{0:J})
\right].
\label{eq:app_agent_pos}
\end{equation}
For consecutive-step displacements
$\mathbf{v}_j=\mathbf{p}_j-\mathbf{p}_{j-1}$ and
$\hat{\mathbf{v}}_j=\hat{\mathbf{p}}_j-\hat{\mathbf{p}}_{j-1}$, the direction
term is
\begin{equation}
\mathcal{L}_{\mathrm{cos}}
=\frac{1}{J}\sum_{j=1}^{J}w_j
\left(1-\frac{\hat{\mathbf{v}}_j^\top\mathbf{v}_j}
{\max(\|\hat{\mathbf{v}}_j\|_2,\epsilon)
 \max(\|\mathbf{v}_j\|_2,\epsilon)}\right),\qquad
w_j=\min\left(\frac{\|\mathbf{v}_j\|_2}{\eta},1\right).
\label{eq:app_agent_dir}
\end{equation}
Here $\eta=0.25$\,m downweights nearly stationary targets, and
$\epsilon=10^{-6}$ stabilizes normalization. Position and direction losses
are averaged over valid annotated tracks; samples labeled \emph{none}
contribute only to classification. The direction loss uses consecutive-step
displacements, whereas position regression uses current-relative offsets.

\paragraph{Meta actions and textual CoT.}
Longitudinal and lateral meta-action predictions use the cross-entropy
objectives defined in Sec.~\ref{sec:method} of the main paper. Textual CoT is supervised by
autoregressive cross-entropy $\mathcal{L}_{\mathrm{cot}}$ on the annotated
response tokens. Each geometry, motion, or maneuver loss uses only samples
labeled for that task.

\paragraph{Trajectory and score supervision.}
Trajectory regression uses the stage-discounted winner-takes-all objective in Sec.~\ref{sec:method} of the main paper. Following iPad~\citep{ipad},
we obtain scoring targets by simulating each final-stage proposal in its
corresponding NAVSIM scene~\citep{navsim}:
\begin{equation}
q_r=\mathrm{NC}_r\,\mathrm{DAC}_r\,
\frac{5\,\mathrm{EP}_r+5\,\mathrm{TTC}_r+2\,\mathrm{Comf}_r}{12}.
\label{eq:app_score_target}
\end{equation}
Metric definitions are given in Table~\ref{tab:app_navsim_metrics}.
For final proposal logits $a_r$, binary cross-entropy uses these
simulator-derived scores $q_r\in[0,1]$ as soft targets:
\begin{equation}
\mathcal{L}_{\mathrm{score}}
=-\frac{1}{R}\sum_{r=1}^{R}
\left[q_r\log\sigma(a_r)+(1-q_r)\log(1-\sigma(a_r))\right],
\label{eq:app_score}
\end{equation}
where $\sigma$ is the sigmoid function.

\paragraph{Auxiliary supervision.}
The expert's auxiliary agent loss $\mathcal{L}_{\mathrm{agent}}^{\mathrm{aux}}$
combines category cross-entropy and the position terms in
Eq.~\eqref{eq:app_agent_pos}, excluding the direction term. Its auxiliary
meta-action loss $\mathcal{L}_{\mathrm{meta}}^{\mathrm{aux}}$ uses the same
maneuver labels as the backbone classifiers. Both objectives supervise the
expert's final refined states.

\subsection{Training Schedule}
\label{sec:app_training}
Table~\ref{tab:train_recipe} summarizes the reference four-phase configuration.
The geometry encoder remains frozen throughout training, while the VLM receives
language and latent-CoT supervision during joint action training.
Outside Qwen and the action expert, the \emph{Heads} group comprises the
trainable adapters, latent-token modules, and prediction heads.

\begin{table}[!htbp]
\supplementtableformat
\setlength{\tabcolsep}{3pt}
\caption{\textbf{Four-phase training configuration.} Batch/GPU and Global denote
per-GPU and total batch sizes, with gradient accumulation of one.
A dash marks an inactive action expert.}
\label{tab:train_recipe}
\begin{tabular}{lccccrrr}
\toprule
& & \multicolumn{3}{c}{Learning rate}
& \multicolumn{3}{c}{Batch configuration} \\
\cmidrule(lr){3-5}\cmidrule(lr){6-8}
Phase & Epochs & Qwen & Action expert & Heads & Batch/GPU & GPUs & Global \\
\midrule
Alignment & $3$ & frozen & -- & $10^{-4}$ & $2$ & $32$ & $64$ \\
Forecasting & $5$ & frozen & -- & $10^{-4}$ & $4$ & $80$ & $320$ \\
Chain & $10$ & $5{\times}10^{-5}$ & -- & $10^{-4}$ & $3$ & $80$ & $240$ \\
Action & $30$ & $2{\times}10^{-5}$ & $10^{-4}$ & $10^{-4}$ & $2$ & $64$ & $128$ \\
\bottomrule
\end{tabular}
\end{table}

\subsection{NAVSIM-WM-CoT}
\label{sec:app_chain}

\paragraph{Annotation structure.}
We construct route-grounded supervision for NAVSIM using future $3$D
annotations and HD-map information.
NAVSIM-WM-CoT associates observed scenes with future depth, key-agent tracks,
longitudinal and lateral meta actions, and textual reasoning annotations.
Table~\ref{tab:data_recipe} summarizes the temporal object-QA and chain
annotation sets before phase-specific sampling. Future observations are used
only to construct supervision targets.

\begin{table}[!htbp]
\supplementtableformat
\setlength{\tabcolsep}{6pt}
\caption{\textbf{NAVSIM-WM-CoT annotation sets.} Counts precede
phase-specific sampling.}

\label{tab:data_recipe}
\begin{tabular}{lrr P{5.6cm}}
\toprule
Annotation set & Train & Validation & Supervision \\
\midrule
Temporal object QA & $176{,}506$ & $25{,}316$ & Language answers, future geometry, and agent motion \\
Chain & $85{,}070$ & $18{,}179$ & Key agent, track, meta actions, and textual rationale \\
\bottomrule
\end{tabular}
\end{table}

\paragraph{Key-agent annotations.}
For each scene, we select at most one key agent from front-camera-visible
objects with complete future tracks and verified spatiotemporal interaction
with the ego route. Agent labels describe realized behavior based on future
ground truth and map-based spatial relations. The \texttt{decision\_causal}
flag identifies agents whose behavior provides evidence supporting the
recorded ego action. The selected agent is labeled as
\emph{vehicle}, \emph{pedestrian}, \emph{cyclist}, or \emph{other};
\emph{none} denotes the absence of a selected agent. A valid motion target
comprises the current position and eight future positions sampled at
$0.5$\,s intervals in the current ego frame. Future offsets follow the
coordinate convention in Appendix~\ref{sec:app_architecture}. Regression
requires a valid track, while category supervision includes \emph{none}.
The validation annotations contain $5{,}087$ scenes with a selected agent and
$13{,}092$ without one. The \emph{none} label refers only to agent selection
and does not exclude other planning constraints.

\paragraph{Meta-action labels.}
Lateral meta-action labels are refined using the original decision,
navigation command, route geometry, and future ego trajectory, while
preserving the original longitudinal label and eight-point ego trajectory.
Longitudinal labels are \emph{maintain speed}, \emph{accelerate},
\emph{decelerate}, \emph{stop}, and \emph{reverse}. Lateral labels are
\emph{keep lane}, \emph{turn left}, \emph{turn right}, \emph{change left},
\emph{change right}, \emph{nudge left}, and \emph{nudge right}. These labels
supervise both the backbone meta-action tokens and the expert's auxiliary
meta-action readouts.

\paragraph{Textual CoT.}
We generated initial textual CoT annotations from templates using route-control
information, the ego-action rationale, and the selected key agent when present.
Other agents provided scene context. GPT checked the drafts against the
structured scene annotations and proposed corrections. Human annotators
reviewed a sample of the revised annotations.

\subsection{Evaluation Settings}
\label{sec:app_eval}

\paragraph{NAVSIM planning metrics.}
\label{sec:app_navsim_eval}
Planning is evaluated using PDMS on NAVSIM~v1 and EPDMS on
NAVSIM~v2~\citep{navsim,navsimv2}. Table~\ref{tab:app_navsim_metrics}
defines the metric abbreviations.

\begin{table}[!htbp]
\supplementtableformat
\setlength{\tabcolsep}{7pt}
\caption{\textbf{NAVSIM metric abbreviations.} Definitions follow the benchmark
protocols~\citep{navsim,navsimv2}. Higher is better.}
\label{tab:app_navsim_metrics}
\begin{tabular}{lll}
\toprule
Abbreviation & Metric & Reported in \\
\midrule
NC & No at-fault collisions & v1, v2 \\
DAC & Drivable-area compliance & v1, v2 \\
TTC & Time-to-collision within bound & v1, v2 \\
EP & Ego progress & v1, v2 \\
Comf. & Comfort & v1 \\
DDC & Driving-direction compliance & v2 \\
TLC & Traffic-light compliance & v2 \\
LK & Lane keeping & v2 \\
HC & History comfort & v2 \\
EC & Extended comfort & v2 \\
\bottomrule
\end{tabular}
\end{table}

We report scores on a $0$--$100$ scale. NAVSIM~v1 evaluates four-second ego
rollouts against non-reactive traffic replayed from logs. PDMS combines collision and drivable-area terms with
weighted progress, time-to-collision, and comfort scores, as in
Eq.~\eqref{eq:app_score_target}. NAVSIM~v2 adds further compliance and
comfort criteria.
We compute per-scene scores before dataset-level aggregation.

\paragraph{LingoQA.}
We use Lingo-Judge accuracy under the LingoQA evaluation
protocol~\citep{lingoqa}. The fine-tuned Qwen3-VL baseline and GeoVLDrive
share the training data and schedule specified in the main comparison.

\paragraph{Intermediate prediction quality.}
We evaluate future depth, key-agent motion, and meta action decoded from the
final model's latent states on the NAVSIM-WM-CoT validation split listed in
Table~\ref{tab:data_recipe}.
In Table~\ref{tab:intermediate_predictions}, category accuracy evaluates the key-agent type.
Trajectory ADE averages Euclidean position errors across future prediction
steps, and FDE measures the endpoint error at $4$\,s. Joint meta-action
accuracy requires both longitudinal and lateral predictions to be correct.
For each sample and future frame, we align the predicted depth to the target
as $d'_{\mathrm{pred}}=d_{\mathrm{pred}}\,\bar d_{\mathrm{gt}}/\bar d_{\mathrm{pred}}$,
where both means use the same valid depth pixels.
Table~\ref{tab:future_depth_val} reports absolute relative error (AbsRel) and
the fraction of valid pixels satisfying
$\max(d'_{\mathrm{pred}}/d_{\mathrm{gt}},\,d_{\mathrm{gt}}/d'_{\mathrm{pred}})<1.25$.

\begin{table}[!htbp]
\supplementtableformat
\setlength{\tabcolsep}{5pt}
\caption{\textbf{Key-agent motion and meta-action predictions on the NAVSIM-WM-CoT validation split.}
Category and meta-action accuracies are percentages. Trajectory errors are in meters.}
\label{tab:intermediate_predictions}
\small
\setlength{\tabcolsep}{6pt}
\begin{tabular}{@{}cccccc@{}}
\toprule
\multicolumn{3}{c}{Key-agent motion}
 & \multicolumn{3}{c}{Meta action} \\
\cmidrule(lr){1-3}\cmidrule(lr){4-6}
\shortstack{Category\\acc. (\%) $\uparrow$}
 & \shortstack{ADE\\(m) $\downarrow$}
 & \shortstack{FDE@4s\\(m) $\downarrow$}
 & \shortstack{Longitudinal\\acc. (\%) $\uparrow$}
 & \shortstack{Lateral\\acc. (\%) $\uparrow$}
 & \shortstack{Joint\\acc. (\%) $\uparrow$} \\
\midrule
$98.4$ & $0.12$ & $0.17$ & $91.5$ & $96.2$ & $90.1$ \\
\bottomrule
\end{tabular}
\end{table}

\begin{table}[!htbp]
\supplementtableformat
\caption{\textbf{Scale-aligned future depth on the NAVSIM-WM-CoT validation split.}
Each horizon has equal weight in the mean.}
\label{tab:future_depth_val}
\small
\setlength{\tabcolsep}{8pt}
\begin{tabular}{@{}lccccc@{}}
\toprule
Metric & $0.5$\,s & $1.0$\,s & $1.5$\,s & $2.0$\,s & Mean \\
\midrule
AbsRel $\downarrow$ & 0.0577 & 0.0607 & 0.0638 & 0.0678 & 0.0625 \\
$\delta < 1.25$ $\uparrow$ & 0.9456 & 0.9425 & 0.9394 & 0.9351 & 0.9407 \\
\bottomrule
\end{tabular}
\end{table}

\begin{table}[!htbp]
\supplementtableformat
\caption{\textbf{Object category accuracy on NAVSIM 3DQA validation questions (\%).}
$^\ast$ marks the fine-tuned baseline.}
\label{tab:3dqa_category_acc}
\small
\setlength{\tabcolsep}{7pt}
\begin{tabular}{lccccc}
\toprule
Model & Vehicle & \shortstack{Vulnerable\\road user} &
\shortstack{Traffic control/\\work zone} & \shortstack{Other static\\object} & Overall \\
\midrule
Qwen3-VL-2B$^\ast$ & 99.79 & 97.89 & 89.94 & \textbf{93.06} & 97.87 \\
\ourrow \textbf{GeoVLDrive} & \textbf{99.83} & \textbf{98.08} & \textbf{90.81}
 & \textbf{93.06} & \textbf{98.00} \\
\bottomrule
\end{tabular}
\end{table}

\begin{table}[!htbp]
\supplementtableformat
\caption{\textbf{Motion accuracy on NAVSIM 3DQA validation questions (\%).}
Overall includes all motion questions.}
\label{tab:3dqa_motion_acc}
\small
\setlength{\tabcolsep}{5pt}
\begin{tabular}{lccccccc}
\toprule
Model & Stationary & Same-direction & Oncoming & Crossing & Cut-in & Cut-out & Overall \\
\midrule
Qwen3-VL-2B$^\ast$ & 91.23 & 83.87 & 73.17 & 63.91 & 0.00 & 0.00 & 85.77 \\
\ourrow \textbf{GeoVLDrive} & \textbf{96.08} & \textbf{90.00} & \textbf{78.10}
 & \textbf{66.67} & \textbf{40.00} & \textbf{25.81} & \textbf{90.99} \\
\bottomrule
\end{tabular}
\end{table}

\FloatBarrier
\section{Reasoning and Planning Visualizations}
\label{sec:app_qual}

\subsection{Latent CoT, Textual CoT, and Action Predictions}
\label{sec:app_qual_chain}
\label{sec:app_qual_depth}
\begin{figure}[H]
\centering
\includegraphics[width=.90\linewidth]{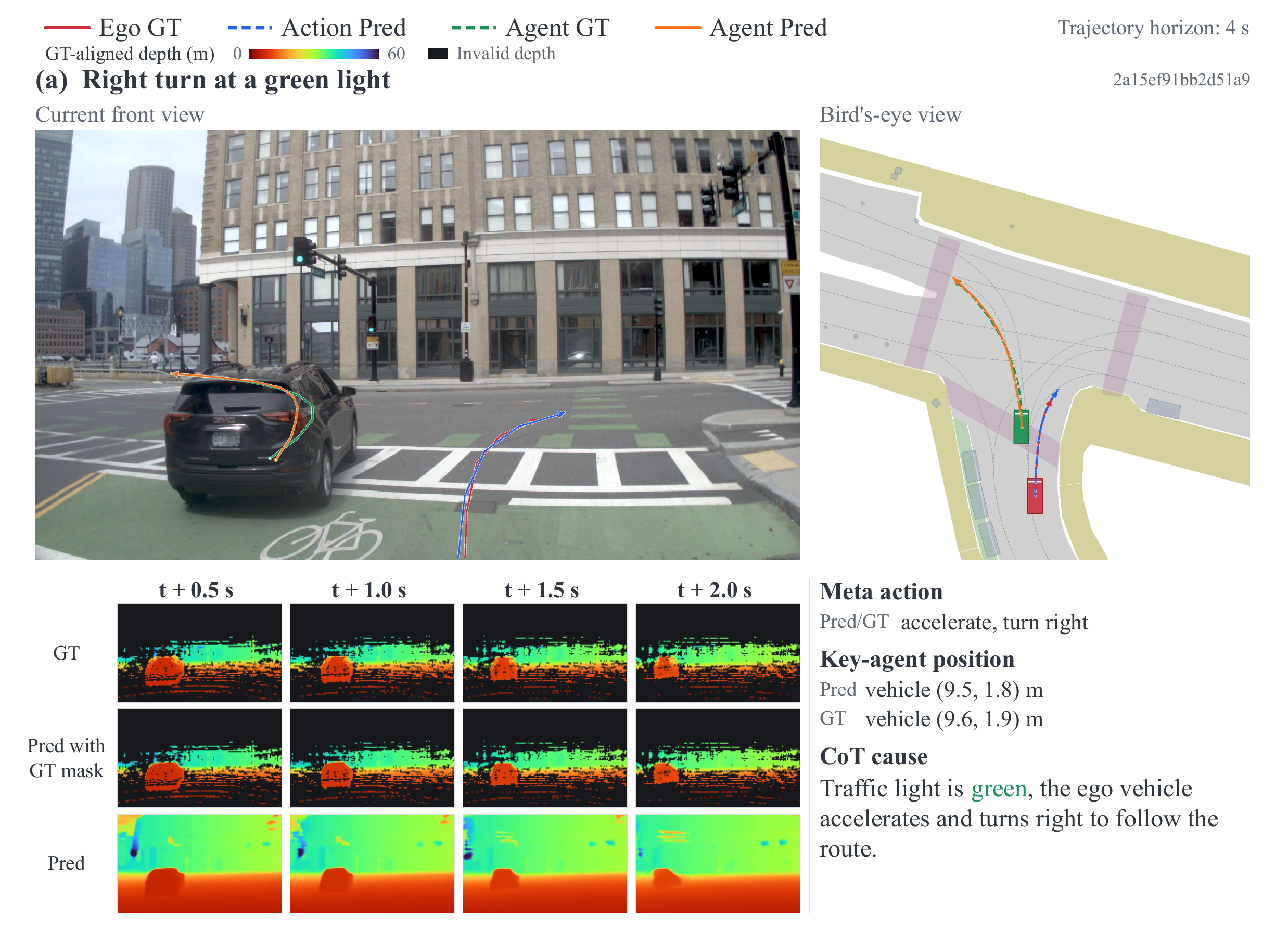}
\caption{\textbf{GeoVLDrive inference on NAVSIM \emph{val}.} Panel (a)
of six selected scenes. Depth rows show
ground truth (GT), predictions at GT-valid pixels, and full predictions.
Depth horizons span $0.5$--$2.0$\,s; trajectories cover $4$\,s. Both prediction
rows are normalized and rescaled to the mean valid GT depth for display.}

\label{fig:app_qual_scenes}
\end{figure}

Figure~\ref{fig:app_qual_scenes} presents model inference on six selected
scenes from the NAVSIM \emph{val} split. The language model generates the
displayed text for illustration. The motion head separately predicts key-agent
positions from the latent states, so the two distance estimates can differ.
Future depth and meta actions are also decoded from latent states. The action
expert predicts the ego trajectory. Positions and trajectories use the current
ego frame.

\begin{samepage}
For visualization, each depth prediction is normalized by its mean
magnitude over the displayed valid pixels and rescaled to the mean valid GT
depth. The masked row uses GT-valid pixels; the full prediction row uses
pixels with finite, positive predicted depth.\par
\end{samepage}

\begin{figure}[H]
\centering
\includegraphics[width=.90\linewidth]{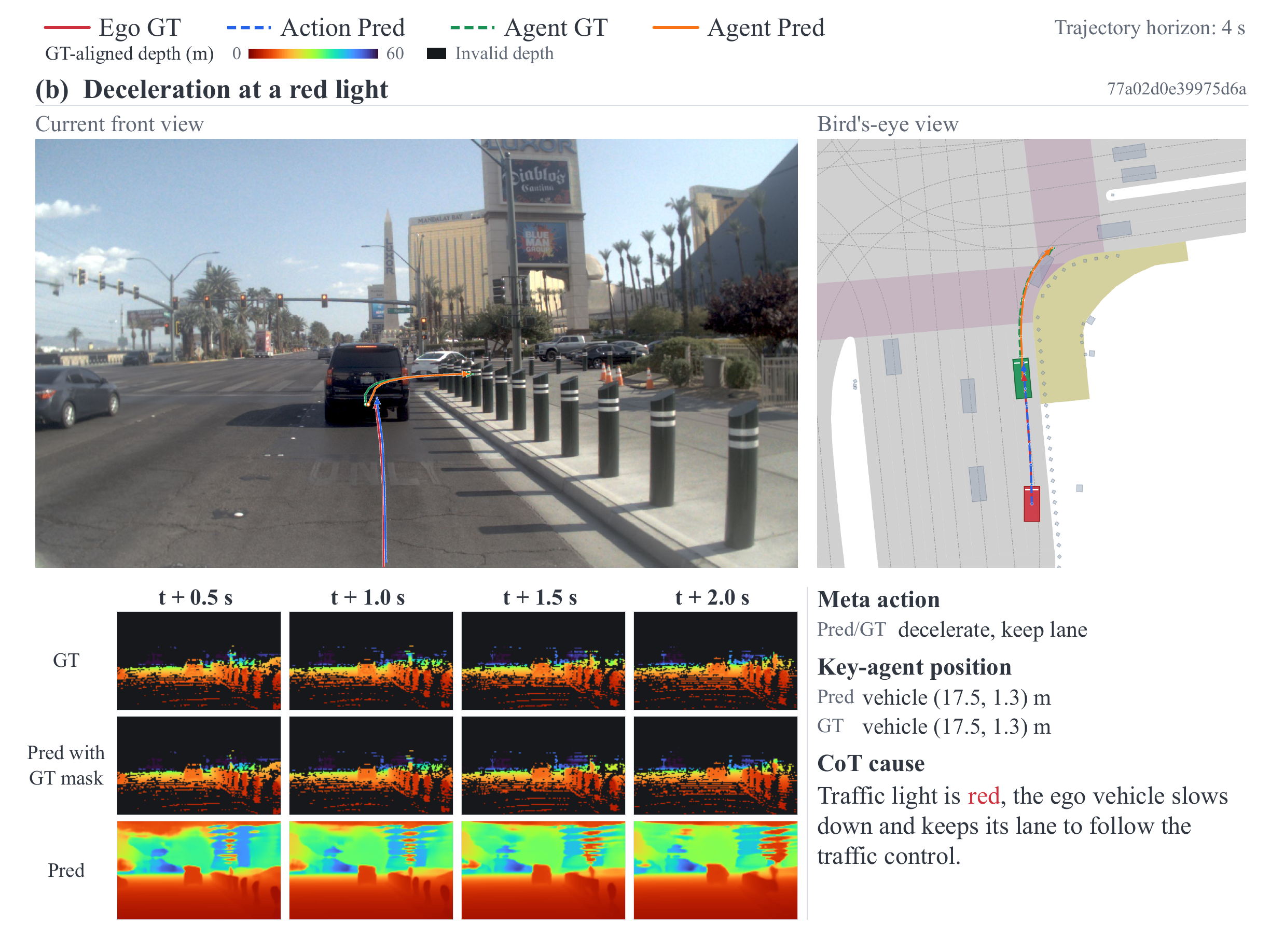}
\par\smallskip
\includegraphics[width=.90\linewidth]{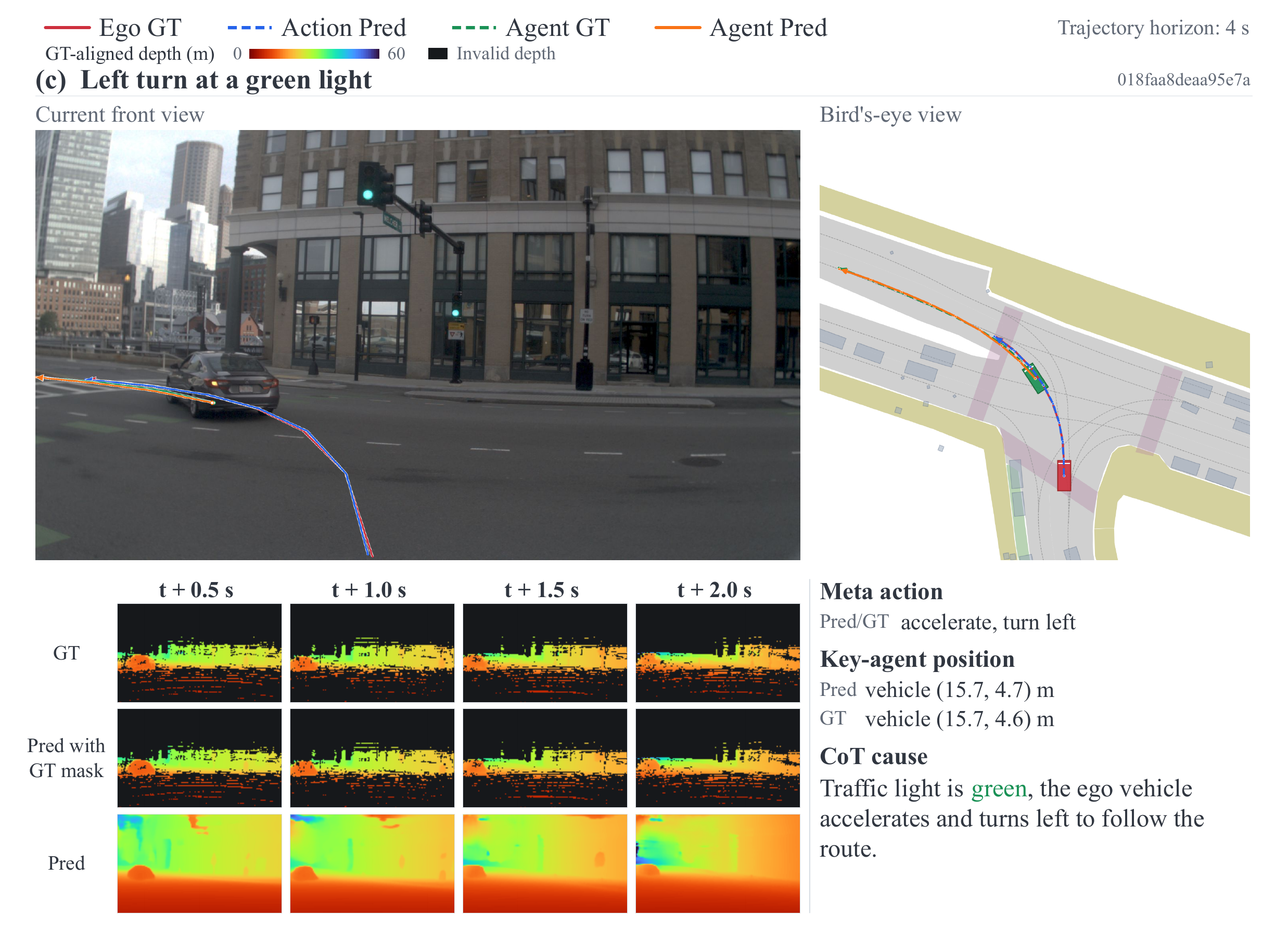}
\continuedfigurecaption{fig:app_qual_scenes}{Panels (b)--(c).}
\end{figure}

\begin{figure}[H]
\centering
\includegraphics[width=.90\linewidth]{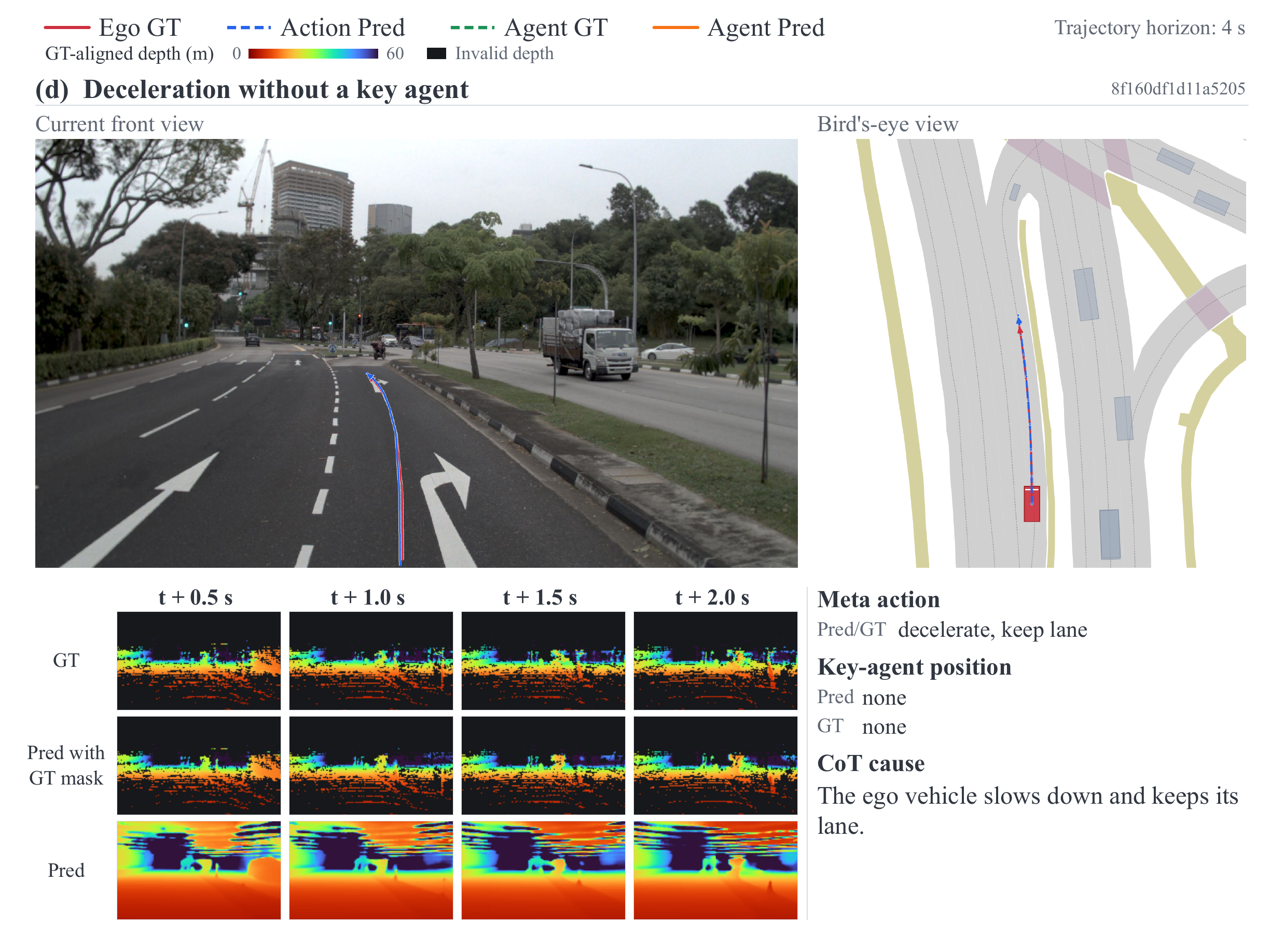}
\par\smallskip
\includegraphics[width=.90\linewidth]{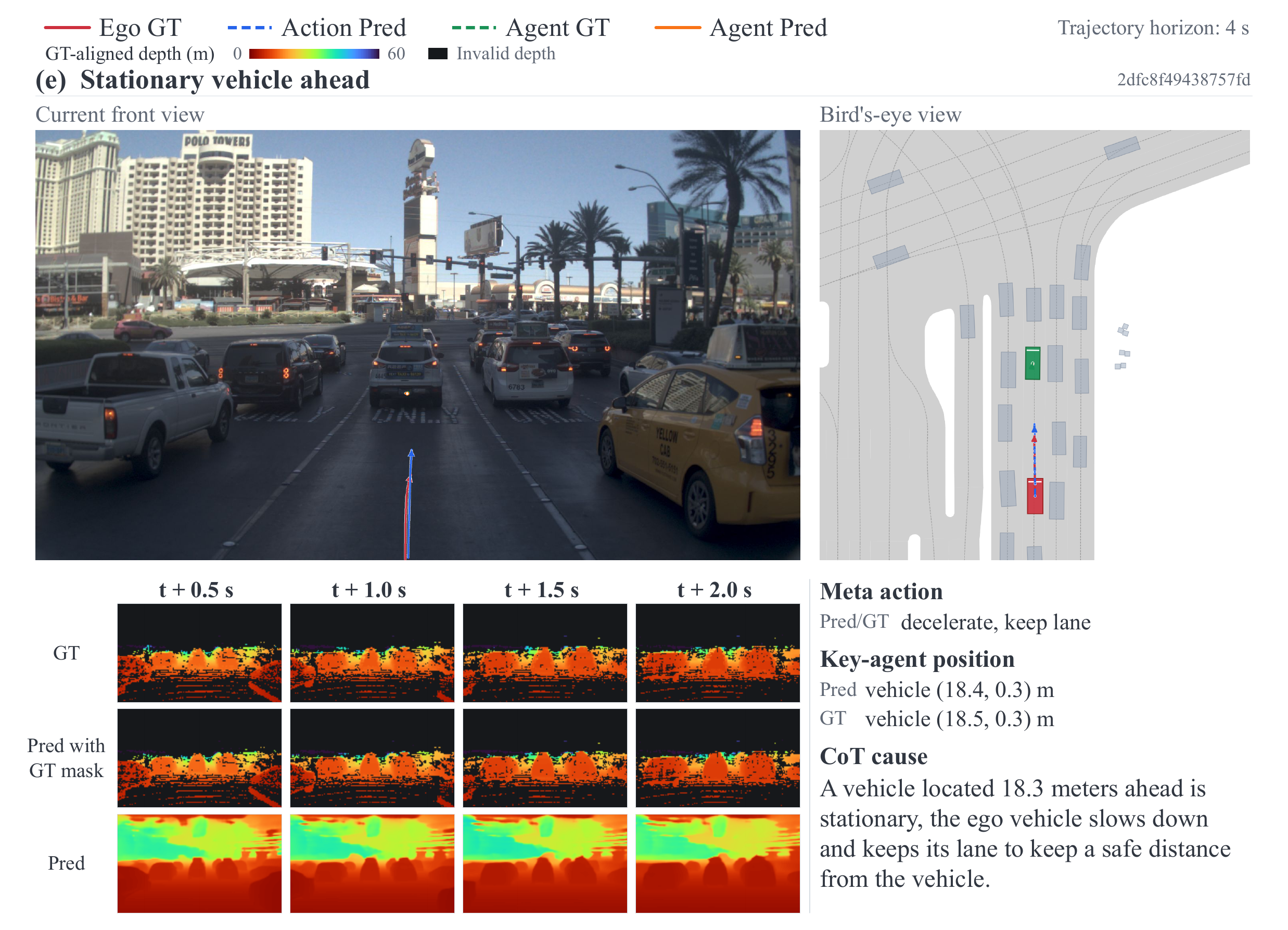}
\continuedfigurecaption{fig:app_qual_scenes}{Panels (d)--(e).}
\end{figure}

\begin{figure}[H]
\centering
\includegraphics[width=.90\linewidth]{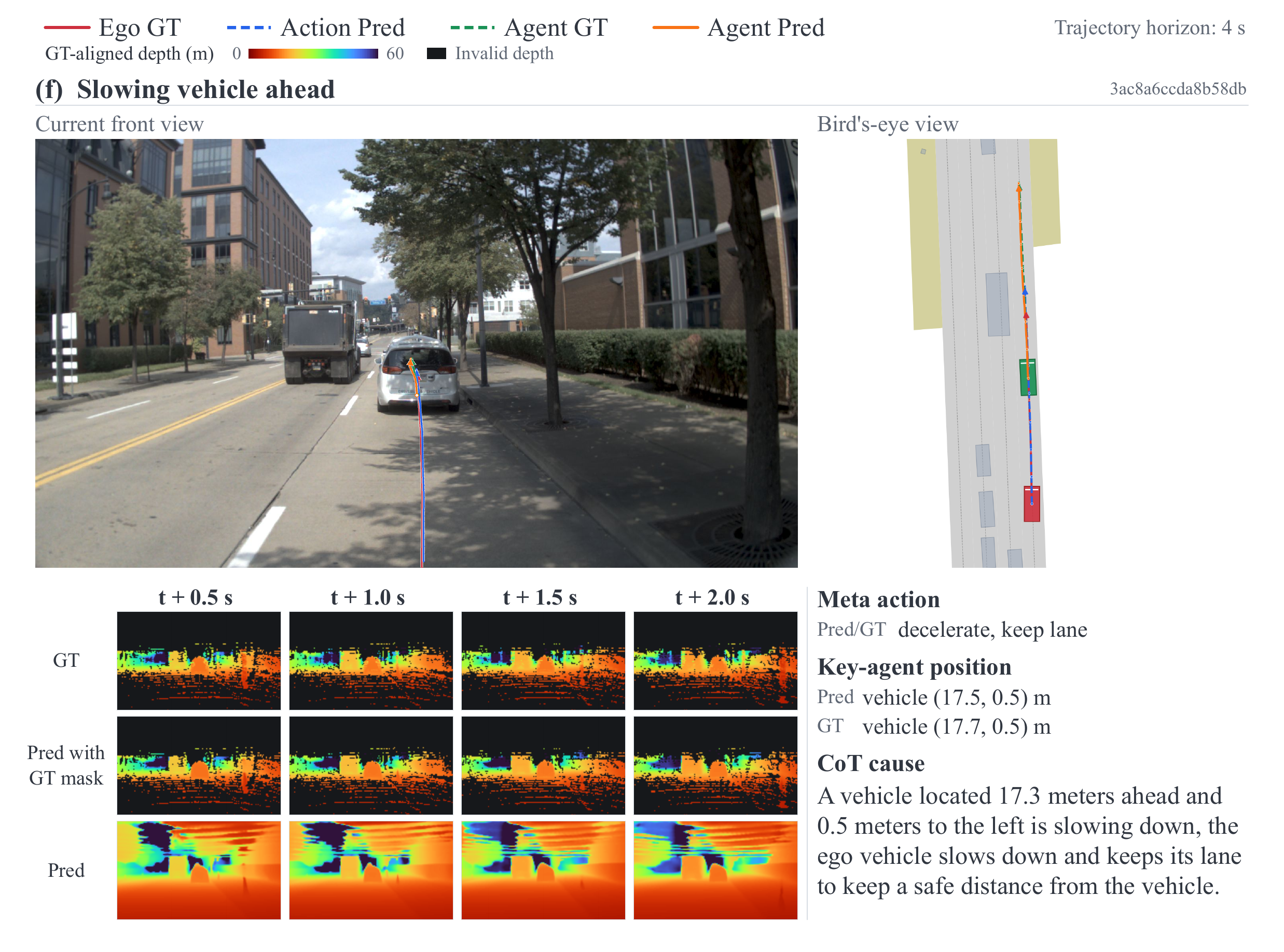}
\continuedfigurecaption{fig:app_qual_scenes}{Panel (f).}
\end{figure}

\FloatBarrier
\subsection{Representative Planning Cases}
\label{sec:app_qual_cases}
\begin{figure}[H]
\centering
\setlength{\parskip}{0pt}
\textbf{(a) Collision avoidance.}\par
\includegraphics[width=0.94\linewidth]{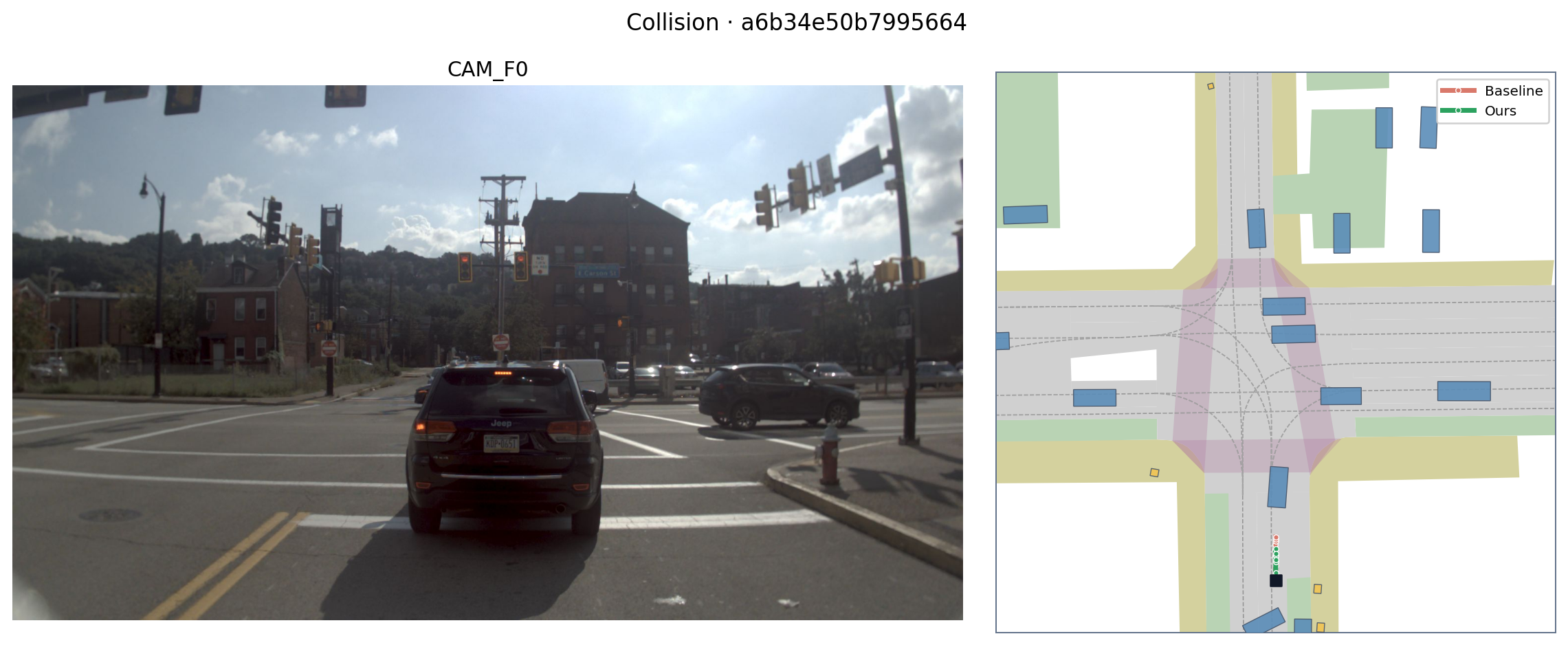}
\caption{\textbf{Qualitative planning comparisons.} Panel (a) of six
selected examples. Each panel pairs a front-camera observation with a
bird's-eye view of GeoWorldAD (red) and GeoVLDrive (green) trajectories.}

\label{fig:app_qual_cases}
\end{figure}

Figure~\ref{fig:app_qual_cases} compares planned trajectories from GeoVLDrive
and GeoWorldAD~\citep{geoworldad} in six selected scenes, grouped by the metric each scene
illustrates: collision avoidance, drivable-area compliance, time-to-collision,
and ego progress.

\begin{figure}[H]
\centering
\setlength{\parskip}{0pt}
\textbf{(b) Collision avoidance.}\par
\includegraphics[width=0.94\linewidth]{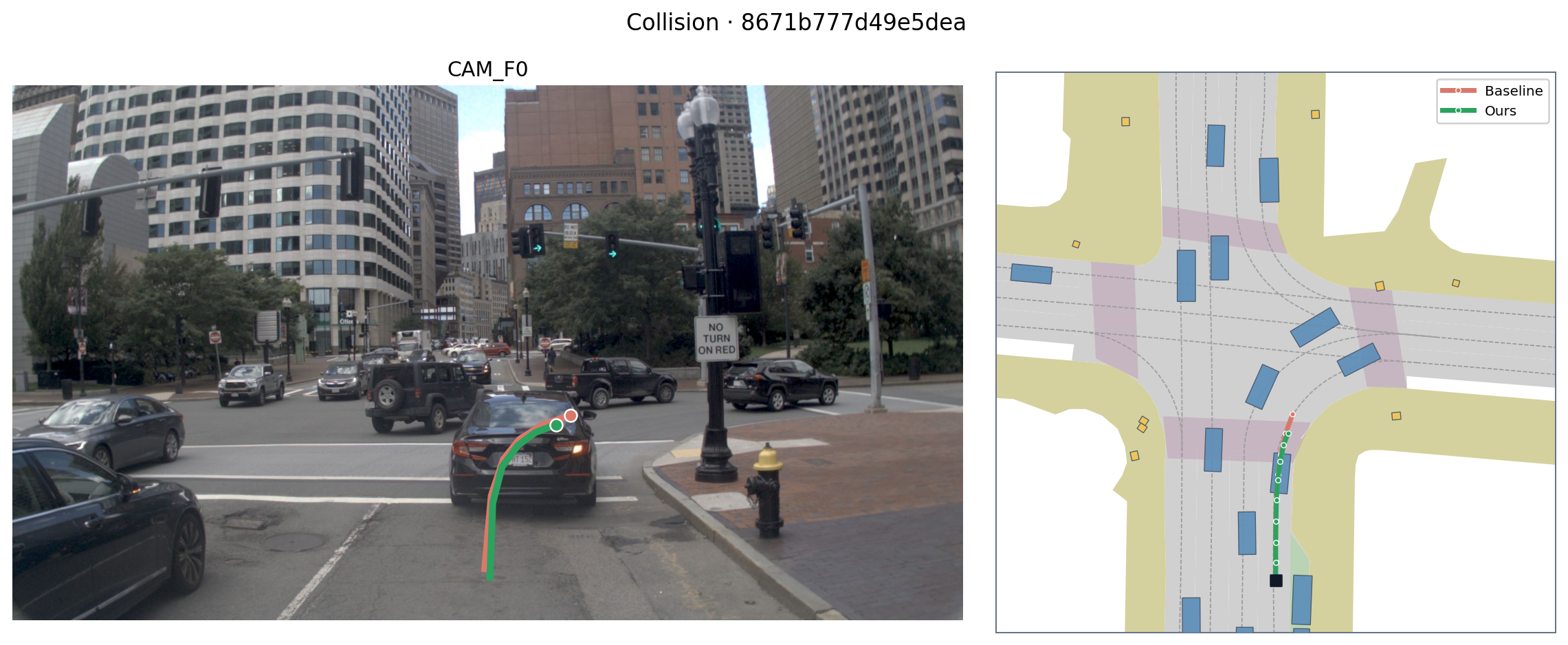}
\par\smallskip
\textbf{(c) Drivable-area compliance.}\par
\includegraphics[width=0.94\linewidth]{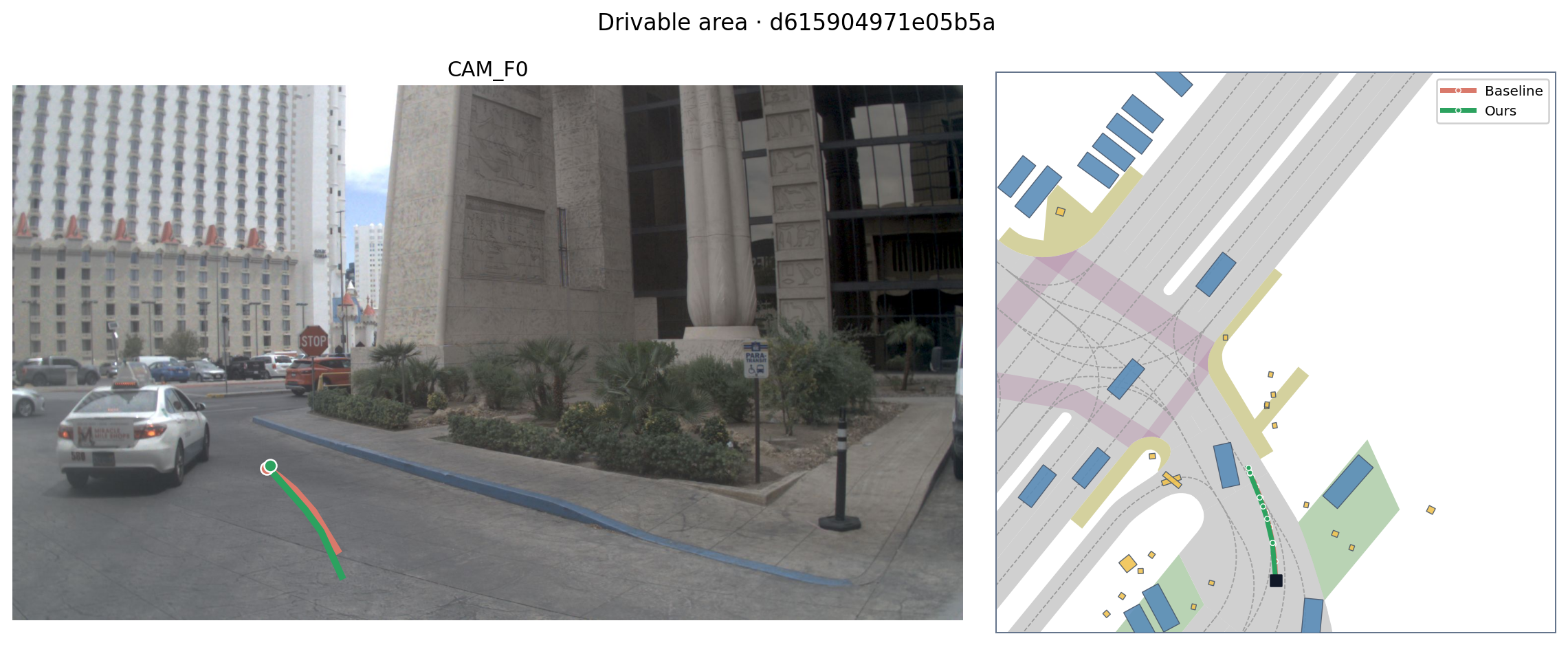}
\par\smallskip
\textbf{(d) Drivable-area compliance.}\par
\includegraphics[width=0.94\linewidth]{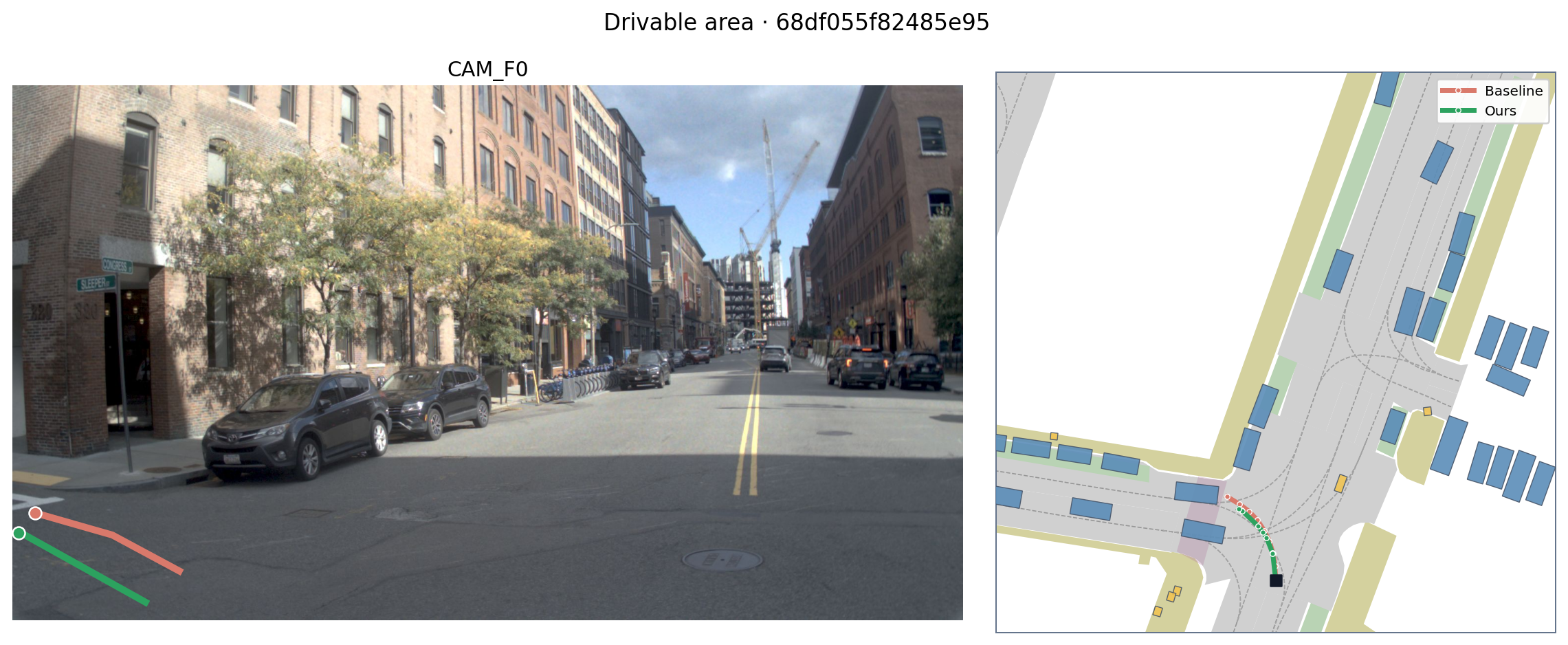}
\continuedfigurecaption{fig:app_qual_cases}{Panels (b)--(d).}
\end{figure}

\begin{figure}[H]
\centering
\setlength{\parskip}{0pt}
\textbf{(e) Time-to-collision.}\par
\includegraphics[width=0.94\linewidth]{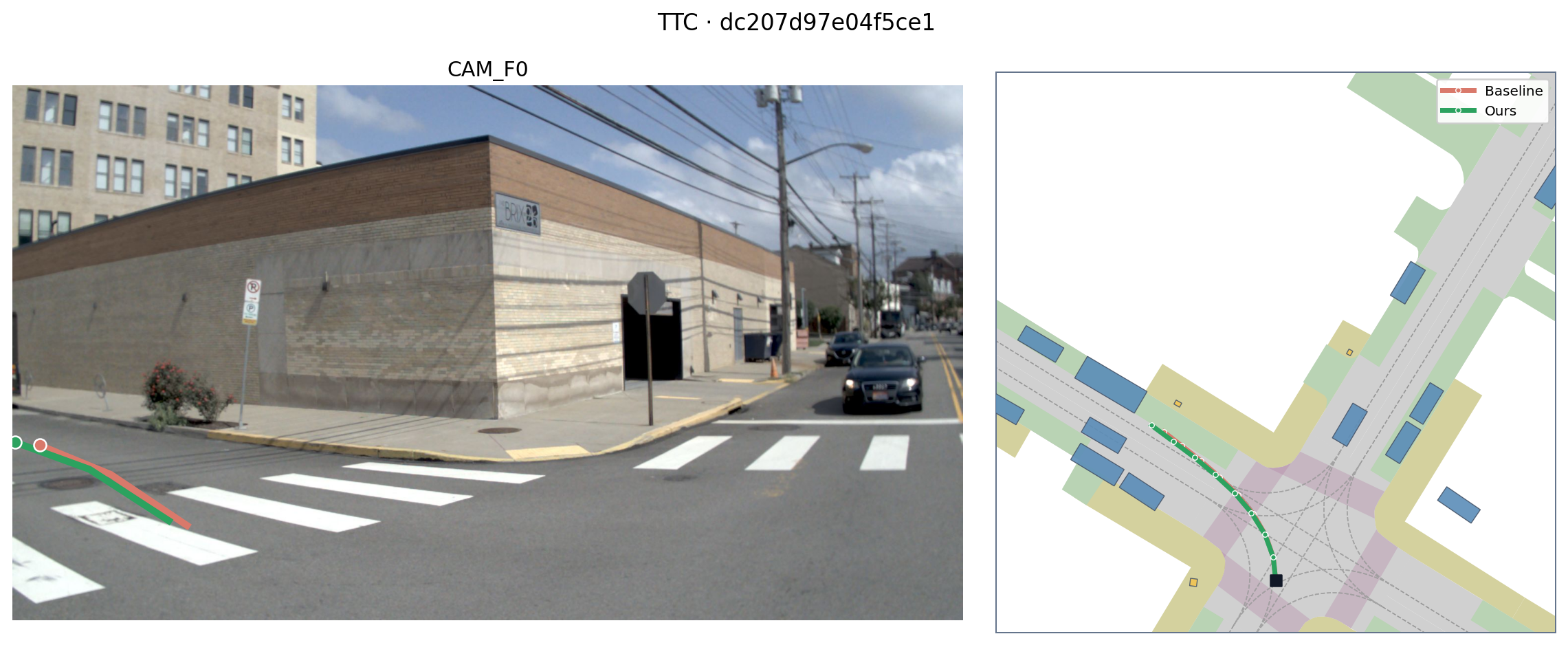}
\par\smallskip
\textbf{(f) Ego progress.}\par
\includegraphics[width=0.94\linewidth]{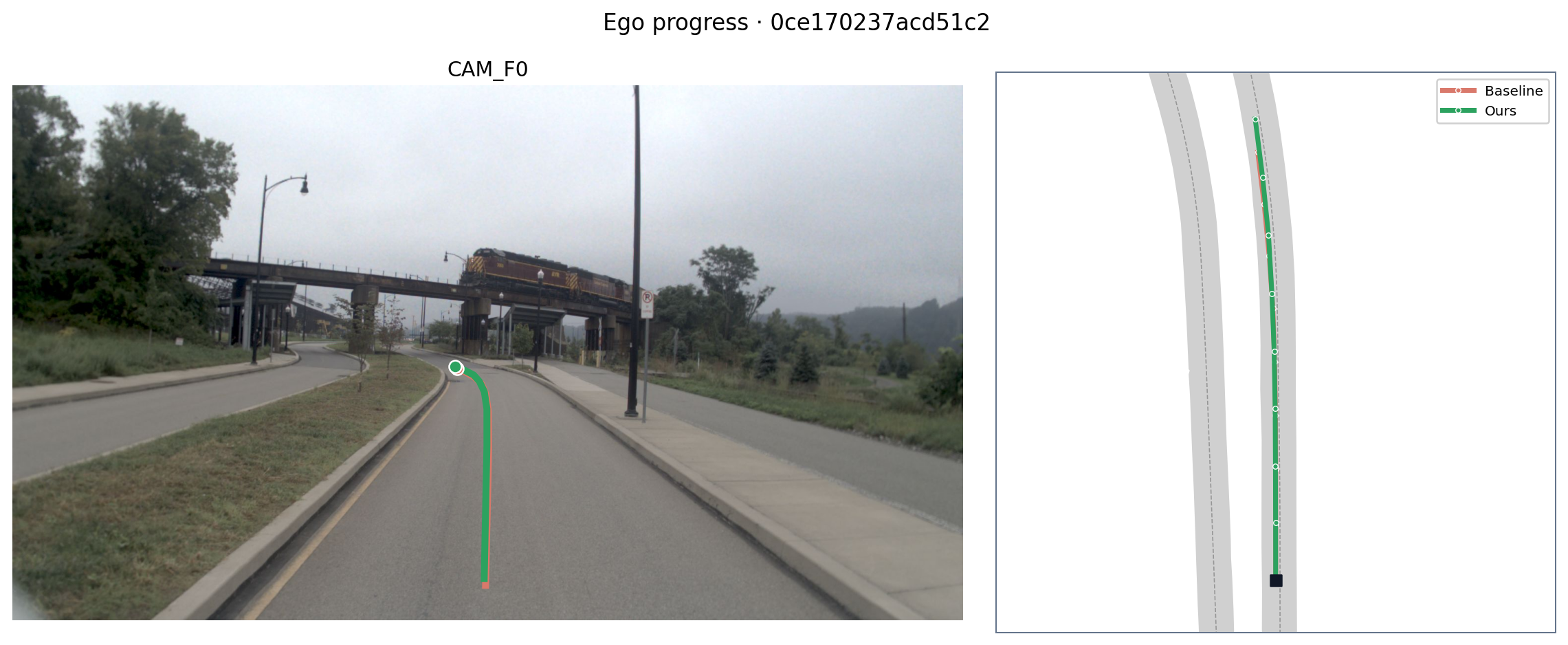}
\continuedfigurecaption{fig:app_qual_cases}{Panels (e)--(f). The color convention is shared with panel (a).}
\end{figure}

\FloatBarrier
\end{document}